\documentclass[10pt,twocolumn,letterpaper]{article}

\usepackage{wacv}
\usepackage{times}
\usepackage{epsfig}
\usepackage{graphicx}
\usepackage{amsmath}
\usepackage{amssymb}

\usepackage{graphicx}
\usepackage{amsmath,amssymb} 
\usepackage{color}
\usepackage{enumitem}
\usepackage{rotating}
\usepackage{times}
\usepackage{epsfig}
\usepackage{graphicx}
\usepackage{amsmath}
\usepackage{amssymb}
\usepackage{amsfonts}
\usepackage{empheq}
\usepackage{framed, color}
\usepackage{xcolor}
\usepackage{graphics}
\usepackage{graphicx}
\usepackage[]{graphicx} 
\usepackage{epsfig} 
\usepackage{subfigure}
\usepackage{algorithm}
\usepackage{algorithmic}
\usepackage{stmaryrd}
\usepackage[mathscr]{eucal}
\usepackage{lineno}
\usepackage{color}
\usepackage{filecontents}
\usepackage{subfigure}
\usepackage{multirow}
\usepackage{filecontents}
\usepackage{verbatim} 
\usepackage{tabularx}
\usepackage{lineno}
\usepackage{setspace}
\usepackage{multirow}
\usepackage{textcomp,booktabs}
\usepackage{caption}
\usepackage{hyperref}

\def\D{{\bf D}}

\def\G{{\bf G}}
\def\g{{\bf g}}
\def\I{{\bf I}}
\def\R{{\bf R}}
\def\X{{\bf X}}
\def\Y{{\bf Y}}

\def\Z{{\bf Z}}
\def\M{{\bf M}}
\def\m{{\bf m}}
\def\n{{\bf n}}

\def\W{{\bf W}}

\def\0{{\bf 0}}
\def\1{{\bf 1}}

\def\alp{\mbox{\boldmath$\alpha$\unboldmath}}

\def\eg{\emph{e.g. }}
\def\ie{\emph{i.e. }}
\def\argmax{\mathop{\rm argmax}}

\newcommand*{\colorboxed}{}
\def\colorboxed#1#{%
  \colorboxedAux{#1}%
}
\newcommand*{\colorboxedAux}[3]{%
  \begingroup
    \colorlet{cb@saved}{.}%
    \color#1{#2}%
    \boxed{%
      \color{cb@saved}%
      #3%
    }%
  \endgroup
}

\wacvfinalcopy 

\def\wacvPaperID{321} 

\ifwacvfinal\fi
\begin{document}

\title{Pixel-wise Attentional Gating for Scene Parsing}

\author{Shu Kong, \ \   Charless Fowlkes \\
Department of Computer Science, \\University of California, Irvine, CA 92697, USA\\
{\tt\small \{skong2, fowlkes\}@ics.uci.edu}\\ \\
\ [\href{http://www.ics.uci.edu/~skong2/PAG.html}{\color{blue}Project Page}]
[\href{https://github.com/aimerykong/Pixel-Attentional-Gating}{\color{blue}Github}],
[\href{http://www.ics.uci.edu/~skong2/slides/20180323_APG_ASU.pdf}{\color{blue}Slides}]
}

\maketitle

\begin{abstract}
To achieve dynamic inference in pixel labeling tasks, we propose
\emph{Pixel-wise Attentional Gating} (\emph{PAG}), which learns to selectively
process a subset of spatial locations at each layer of a deep convolutional
network.  PAG is a generic, architecture-independent, problem-agnostic
mechanism that can be readily ``plugged in'' to an existing model with
fine-tuning.  We utilize PAG in two ways: 1) learning spatially varying pooling
fields that improve model performance without the extra computation cost
associated with multi-scale pooling, and 2) learning a dynamic computation
policy for each pixel to decrease total computation (FLOPs) while maintaining
accuracy.

We extensively evaluate PAG on a variety of per-pixel labeling tasks, including
semantic segmentation, boundary detection, monocular depth and surface normal
estimation.  We demonstrate that PAG allows competitive or state-of-the-art
performance on these tasks.  Our experiments show that PAG learns dynamic
spatial allocation of computation over the input image which provides better
performance trade-offs compared to related approaches (e.g., truncating deep
models or dynamically skipping whole layers). Generally, we observe PAG can
reduce computation by $10\%$ without noticeable loss in accuracy and
performance degrades gracefully when imposing stronger computational constraints.
\end{abstract}

\section{Introduction}
\label{sec:intro}


The development of deep convolutional neural networks (CNN) has allowed remarkable
progress in wide range of image pixel-labeling tasks such as boundary
detection~\cite{maninis2017convolutional,xie2015holistically,kokkinos2015pushing},
semantic segmentation~\cite{kong2017recurrentpixel,kong2017recurrentscene,chen2016deeplab},
monocular depth estimation~\cite{kong2017recurrentscene,li2015depth,laina2016deeper,liu2015deep,eigen2014depth},
and surface normal estimation~\cite{wang2015designing,bansal2016marr,eigen2015predicting}.
Architectures that enable training of increasingly deeper networks have
resulted in corresponding improvements in prediction
accuracy~\cite{simonyan2014very,he2016deep}.  However, with great depth comes
great computational burden. This hinders deployment of such deep models in edge
and mobile computing applications which have significant power/memory constraints.

To make deep models more practically applicable, a flurry of recent work has
focused on reducing these storage and computational costs
\cite{han2015deep,molchanov2016pruning,iandola2016squeezenet,mallya2017packnet,carreira2017model,kong2017low}.
Static offline techniques like network distillation~\cite{hinton2015distilling},
pruning~\cite{molchanov2016pruning}, and model compression~\cite{carreira2017model} take a trained
network as input and synthesize a new network that approximates the same functionality
with reduced memory footprint and test-time execution cost. Our approach is
inspired by a complementary family of techniques that learn to vary the network
computation depth adaptively, depending on the input
data~\cite{veit2017convolutional,wu2017blockdrop,wang2017skipnet,figurnov2017spatially}.

In this paper, we study the problem of achieving dynamic inference for
per-pixel labeling tasks with a deep CNN model under limited computational
budget.
For image classification, dynamic allocation of computational
``attention'' can be interpreted as expending more computation on ambiguous
images (e.g.,~\cite{veit2017convolutional,wu2017blockdrop,wang2017skipnet}) or
limiting processing to informative image regions~\cite{figurnov2017spatially}.
However, understanding the role of dynamic
computation in pixel labeling tasks has not been explored.  Pixel-level
labeling requires analyzing fine-grained image details and making predictions
at every spatial location, so it is not obvious that dynamically allocating
computation to different image regions is useful.  Unlike classification,
labeling locally uninformative regions would seem to demand more computation
rather than less (e.g., to incorporate long-range context).

\begin{figure*}[t]
\centering
   \includegraphics[width=0.99\linewidth]{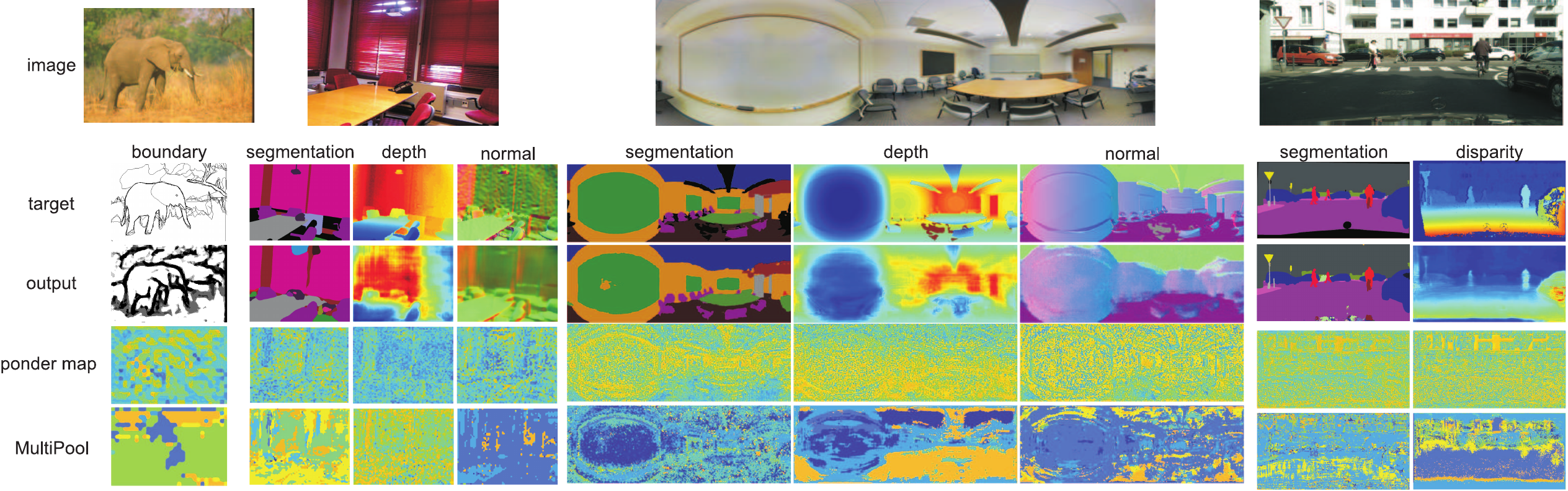}
   \vspace{-3mm}
   \caption{Pixel-wise Attentional Gating units (PAG) achieve dynamic
   inference by learning a dynamic computation path for each pixel under limited
   computation budget. The ``ponder maps'' shown in the last row provide a
   visualization of the amount of computation allocated to each location
   (generated by accumulating binary masks from PAG units across all layers);
   whereas the ``MultiPool'' adaptively chooses the proper pooling size for each
   pixel to aggregate information for inference.
   We apply PAG to a variety of per-pixel labeling tasks (boundary detection,
   semantic segmentation, monocular geometry) and evaluate over diverse
   image datasets (indoor/outdoor scenes, narrow/wide field-of-view).}
\vspace{-2mm}
\label{fig:splashFigure}
\end{figure*}


To explore these questions, we introduce a \emph{Pixel-wise Attentional Gating}
(\emph{PAG}) unit that selects a sparse subset of spatial locations to process
based on the input feature map.  We utilize the Gumbel sampling trick
\cite{gumbel2012statistics,jang2016categorical,maddison2016concrete}
to allow differentiable, end-to-end latent training of PAG units inserted
across multiple computational blocks of a given task-specific architecture.  We
exploit this generic PAG unit in two ways: bypassing sequential (residual)
processing layers and dynamically selecting between multiple parallel network
branches.

\noindent{\em Dynamic computation depth:} Inserting PAG at multiple layers of a Residual
Network enables learning a dynamic, feed-forward computation path for each
pixel that is conditional on the input image (see the second last row in Fig.~\ref{fig:splashFigure}).
We introduce a sparsity
hyperparameter that provides control over the average total and per-layer
computation.  For a fixed computational budget, we show this dynamic,
per-pixel gating outperforms architectures that meet the budget by either using a
smaller number of fixed layers or learning to dynamically bypass whole layers
(Section~\ref{ssec:parsimony}).

\noindent{\em Dynamic spatial pooling:} We exploit PAG to dynamically select the extent
of pooling regions at each spatial image location (see the last row in Fig.~\ref{fig:splashFigure}).
Previous work has
demonstrated the benefits of averaging features from multiple pooling scales
using either learned weights~\cite{chen2016deeplab}, or spatially varying weights based
on attention~\cite{chen2016attention} or scene depth~\cite{kong2017recurrentscene}. However,
such multi-scale pooling requires substantially more computation. We show the
proposed PAG unit can learn to select appropriate spatially-varying pooling,
outperforming the recent work of \cite{kong2017recurrentscene} without the
computational burden of multiple parallel branches (Section~\ref{ssec:multipool}).

We carry out an extensive evaluation of pixel-wise attentional gating over
diverse datasets for a variety of per-pixel labeling tasks including boundary
detection, semantic segmentation, monocular depth estimation and surface normal
estimation (see Fig.~\ref{fig:splashFigure}).  We demonstrate that PAG helps
deliver state-of-the-art performance on these tasks by dynamically allocating
computation.  In general, we observe that the introduction of PAG units can
reduce total computation by $10\%$ without noticeable drop in accuracy and
shows graceful degradation in performance even with substantial budget
constraints (e.g., a $30\%$ budget cut).

To summarize our primary contribution: (1) we introduce a pixel-wise attentional
gating unit which is problem-agnostic, architecture-independent and provides a
simple method to allow user-specified control computational parsimony with
standard training techniques; (2) we investigate the role of dynamic
computation in pixel-labeling tasks and demonstrate improved prediction
performance while maintaining or reducing overall compute cost.

\section{Related Work}

Deep CNN models with residual or ``skip'' connections have yielded substantial
performance improvements with increased
depth~\cite{he2016deep,huang2017densely}, but also introduced redundant
parameters and computation~\cite{han2015deep,molchanov2016pruning}.  In
interpreting the success of residual networks (ResNet)~\cite{he2016deep}, it
has been suggested that ResNet can be seen as an ensemble of many small
networks~\cite{veit2016residual}, each defined by a path through the network
topology. This is supported by the observation that ResNet still performs well
even when some layers are removed after
training~\cite{huang2016deep,figurnov2017spatially}.  This indicates it may be
possible to reduce test-time computation by dynamically choosing only a subset
of these paths to evaluate
\cite{veit2017convolutional,wu2017blockdrop,wang2017skipnet,figurnov2017spatially}.

This can be achieved by learning a halting policy that stops computation
after evaluation of a particular layer~\cite{figurnov2017spatially}, or a more
flexible routing policy trained through reinforcement
learning~\cite{wu2017blockdrop,wang2017skipnet}. Our method is most closely
related to~\cite{veit2017convolutional}, which utilizes the ``Gumbel sampling
trick''~\cite{gumbel2012statistics,jang2016categorical,maddison2016concrete}
to learn binary gating that determines whether each layer is computed.
The Gumbel sampling technique allows one to perform gradient descent on
models that include a discrete argmax operation without resorting to
approximation by softmax or reinforcement learning techniques.

The PerforatedCNN~\cite{figurnov2016perforatedcnns} demonstrated that
convolution operations could be accelerated by learning static masks that skip
computation at a subset of spatial positions.  This was used
in~\cite{figurnov2017spatially} to achieve spatially varying dynamic depth. Our
approach is simpler (it uses a simple sparsity regularization to directly control
amount per-pixel or per-layer computation rather than ponder cost) and more
flexible (allowing more flexible routing policies than early
halting\footnote{Results in~\cite{wu2017blockdrop,wang2017skipnet} suggest general
routing offers better performance than truncating computation at a particular
depth.}).

Finally, our use of dynamic computation to choose between branches is related
to \cite{kong2017recurrentscene}, which improves semantic segmentation by
fusing features from multiple branches with various pooling sizes using a
spatially varying weighted average.  Unlike
\cite{chen2016deeplab,chen2016attention,kong2017recurrentscene}
which require computing the outputs of
parallel pooling branches, our PAG-based learns to select a pooling size for
each spatial location and only computes the necessary pooled features.
This is similar in spirit to the work of~\cite{shazeer2017outrageously}, which
demonstrated that sparsely-gated mixture-of-experts can dramatically increase
model capacity using multi-branch configuration with only minor losses in
computational efficiency.

\section{Pixel-wise Attentional Gating}
\label{sec:PAG}
We first describe our design of Pixel-wise Attentional Gating (PAG) unit and
its relation to the ResNet architecture~\cite{he2016deep}.  Then, we elaborate
how we exploit the Gumbel sampling technique to learning PAG differentiable
even when generating binary masks. Finally we describe how the PAG unit can
be used to perform dynamic inference by (1) selecting the subset of layers in
the computational path for each spatial location, and (2) selecting the correct pooling
size at each spatial location.

\subsection{Plug-in PAG inside a Residual Block} \label{ssec:PAG_in_resBlock}
Consider a block that computes output $\bf O$ using a residual update $\Z={\cal
F}(\I)$ to some input $\I$. To reduce computation, one can learn a gating
function ${\cal G}(\I)$ that selects a subset of spatial locations (pixels) to
process conditional on the input. We represent the output of ${\cal G}$ as a
binary spatial mask $\G$ which is replicated along feature channel dimension as
needed to match dimension of ${\bf O}$ and $\I$. The spatially gated residual
update can be written as:
\begin{equation}
\small
\begin{split}
\G=& {\cal G}(\I) \\
{\bf O}=&{\bar \G}\odot\I + \G \odot ({\cal F}_{\G}( \I) + \I) \\
 =&\I + \G \odot {\cal F}_{\G}( \I)
\end{split}
\label{eq:gating_in_layers}
\end{equation}
where $\odot$ is element-wise product, ${\bar \G} = 1-{\G}$, and the notation
${\cal F}_{\G}$ indicates that we only evaluate ${\cal F}$ at the locations
specified by $\G$.  An alternative to spatially varying computation is for the
gating function to predict a single binary value that determines whether or not
the residual is calculated at this layer~\cite{veit2017convolutional} in which
case ${\cal F}_{\G}$ is only computed if $\G=1$.

\begin{figure}[t]
\centering
   \includegraphics[width=0.99\linewidth]{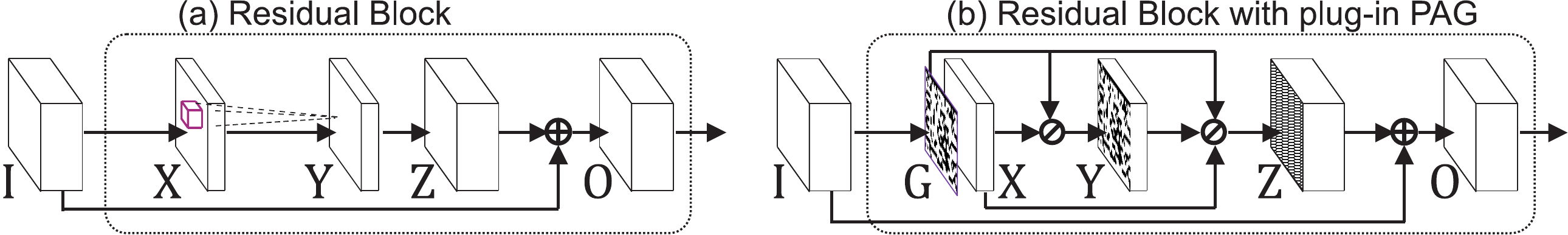}
\vspace{-3mm}
   \caption{\small (a) A standard residual block. (b) Pixel-wise Attentional Gating
   unit (PAG) integrated into a residual block. Boxes/arrows denote
   activations/computations.  $\G$ is a sparse, binary map that modulates what
   processing applied to each spatial location.
   ``$\oslash$'' means the perforated convolution~\cite{figurnov2016perforatedcnns},
   which assembles only active pixels for computation.
   }
\vspace{-4mm}
\label{fig:sketch_module}
\end{figure}

Both pixel-wise and layer-wise gating have the intrinsic limitation that the
gating function ${\cal G}$ must be evaluated prior to ${\cal F}$,
inducing computational delays. To overcome this
limitation we integrate the gating function more carefully within the ResNet
block. We demonstrate ours in the equations below comparing a standard residual
block (left) and the one with PAG (right), respectively with corresponding
illustrations in Fig.~\ref{fig:sketch_module}:
\begin{equation}
\small
\begin{aligned}[c]
\X=& {\cal F}^{1} (\I) \\
\Y = & {\cal F}^{2}(\X) \\
\Z =&{\cal F}^{3}(\Y)\\
{\bf O} = & \I + \Z \\
\end{aligned}
\qquad\qquad
\begin{aligned}[c]
\X = & {\cal F}^{1} (\I), \ \ \G={\cal G}(\I) \\
\Y = & {\cal F}_{\G}^{2}(\X) \\
\Z = & {\cal F}^{3}_{\G}(\bar{\G} \odot \X + \G \odot \Y)\\
{\bf O} = & \I + \Z  \\
\end{aligned}
\label{eq:PAG_in_resblock}
\end{equation}

The transformation functions ${\cal F}$'s consist of convolution, batch
normalization~\cite{ioffe2015batch} and ReLU~\cite{nair2010rectified} layers.
As seen from the right set of equations, our design advocates computing the
gating mask on the input $\I$ to the current building block in parallel with
$\X={\cal F}_{\X}(\I)$.  ResNet adopts bottleneck structure so the first
transformation ${\cal F}^{1}$ performs dimensionality reduction with a set of
1$\times$1 kernels, ${\cal F}^{2}$ utilizes 3$\times$3 kernels, and ${\cal
F}^{3}$ is another transform with 1$\times$1 kernels that restores
dimensionality.  As a result, the most costly computation is in the second
transformation ${\cal F}^{2}$ which is mitigated by gating the computation.
We show in our ablation study (Section~\ref{ssec:ablation}) that for
per-pixel labeling tasks, this design outperforms layer-wise gating.

\subsection{Learning Discrete Attention Maps}
The key to the proposed PAG is the gating function $\cal G$ that produces a
discrete (binary) mask which allows for reduced computation.  However,
producing the binary mask using hard thresholding is non-differentiable, and
thus cannot be simply incorporated in CNN where gradient descent is used for
training.  To bridge the gap, we exploit the Gumbel-Max
trick~\cite{gumbel2012statistics} and its recent continuous
relaxation~\cite{maddison2016concrete,jang2016categorical}.

A random variable $m$ follows a Gumbel distribution if $m \equiv -\log(-\log(u))$,
where $u$ is a sample from the uniform distribution $u \sim {\cal U}[0,1]$.
Let $g$ be a discrete random variable with probabilities $P(g=k) \propto a_k$,
and let $\{m_k\}_{k=1,\dots,K}$ be a sequence of i.i.d. Gumbel random variables.
Then we can sample from the discrete variable with:
\begin{equation}
\small
\begin{split}
g = & \argmax_{k=1,\dots,K}(\log\alpha_k + m_k)\\
\end{split}
\label{eq:argmax}
\end{equation}
The drawback of this approach is that the argmax operation is not continuous
when mapping the Gumbel samples to the realizations of discrete distribution.
To address this issue, a continuous relaxation the Gumbel Sampling Trick,
proposed in~\cite{maddison2016concrete,jang2016categorical}, replaces the
argmax operation with a softmax. Using a one-hot vector $\g=[g_1,\dots,g_K]$ to
encode $g$, a sample from the Gumbel softmax relaxation can be expressed by the
vector:
\begin{equation}
\small
\begin{split}
\g = & softmax((\log(\alp) + \m)/\tau ) \\
\end{split}
\label{eq:GumbelArgmax}
\end{equation}
where $\alp=[\alpha_1,\dots,\alpha_K]$, $\m=[m_1,\dots,m_K]$, and $\tau$ is the
``temperature'' parameter.  In the limit as $\tau \rightarrow 0$, the softmax
function approaches the argmax function and Eq. (\ref{eq:GumbelArgmax})
becomes equivalent to the discrete sampler.  Since the softmax function is
differentiable and $\m$ contains i.i.d Gumbel random variables which are
independent to input activation $\alp$, we can easily propagate gradients to
the probability vector $\alp$, which is treated as the gating mask for a single
pixel in the per-pixel labeling tasks.

As suggested in~\cite{veit2017convolutional},
we employ the straight-through version~\cite{maddison2016concrete} of
Eq. (\ref{eq:GumbelArgmax}) during training.
In particular,
for the forward pass,
we use discrete samples from Eq. (\ref{eq:argmax}),
but during the backwards pass,
we compute the gradient of the softmax relaxation in Eq. (\ref{eq:GumbelArgmax}).
Based on our empirical observation as well as that reported in~\cite{maddison2016concrete},
such greedy straight-through estimator performs slightly better than strictly following
Eq. (\ref{eq:GumbelArgmax}),
even though there is a mismatch between forward and backward pass.
In our work, we initialize $\tau=1$ and decrease it to $0.1$ gradually during
training.  We find this works even better than training with a constant small
$\tau$.

\subsection{Dynamic Per-Pixel Computation Routing}
\label{ssec:parsimony}
By stacking multiple PAG residual blocks, we can construct a model in which the
subset of layers used to compute an output varies for each spatial location
based the collection of binary masks. We allow the user to specify the
computational budget in terms of a target sparsity $\rho$.  For a binary mask
$\G \in \{0,1\}^{H\times W}$, we compute the empirical sparsity
$g=\frac{1}{H*W}\sum_{h,w}^{H,W} \G_{h,w}$ (smaller values indicate sparser
computation) and measure how well it matches the target $\rho$ using the KL
divergence:
\begin{equation}
\small
\begin{split}
KL(\rho\Vert g) \equiv \rho \log(\frac{\rho}{g}) + (1-\rho)\log(\frac{1-\rho}{1-g})
\end{split}
\label{eq:KL}
\end{equation}

To train the model, we jointly minimize the sum of a task-specific loss
$\ell_{task}$ and the per-layer sparsity loss summed over all layers
of interest:
\begin{equation}
\small
\begin{split}
\ell=\ell_{task} + \lambda \sum_{l=1}^{L} KL(\rho \Vert g_l)
\end{split}
\label{eq:totalLoss}
\end{equation}
where $l$ indexes one of $L$ layers which have PAG inserted for dynamic
computation and $\lambda$ controls the weight for the constraints.  In our
experiments we set $\lambda=10^{-4}$ but found performance is stable over a wide
range of penalties ($\lambda \in [10^{-5},10^{-2}]$). To visualize the spatial
distribution of computation, we accumulate the binary gating masks from all to
produce a ``ponder map''. This reveals that trained models do not allocate
computation uniformly, but instead responds to image content (\eg focusing
computation on boundaries between objects where semantic labels, depths
or surface normals undergo sharp changes).

An alternative to per-layer sparsity is to compute the total sparsity
$g=\frac{1}{L}\sum_{l=1}^L g_l$ and penalize $g$ with $KL(\rho\Vert g)$.
However, training in this way does not effectively learn dynamic computational
paths and results in trivial, non-dynamic solutions, \eg completely skipping a
subset of layers and always using the remaining ones.  Similar phenomenon is
reported in~\cite{veit2017convolutional}.
In training models we typically start
from a pre-trained model and insert sparsity constraints progressively.
We found this incremental construction produces better diversity in the PAG
computation paths.  We also observe that when targeting reduced computation
budget, fine-tuning a model which has already been trained with larger $\rho$
consistently brings better performance than fine-tuning a pre-trained model
directly with a small $\rho$.

\subsection{Dynamic Spatial Pooling}
\label{ssec:multipool}

\begin{figure}[t]
\centering
   \includegraphics[width=0.99\linewidth]{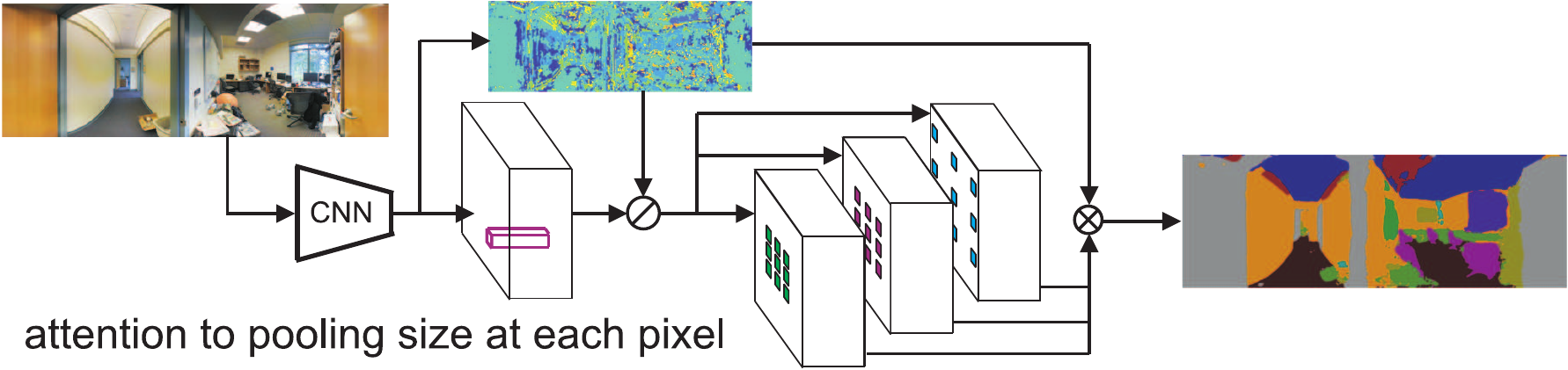}
\vspace{-3mm}
   \caption{\small PAG-based MultiPool module learns to select the pooling size for
   each spatial location so that contextual information can be better
   aggregated. This can be implemented efficiently using perforated
   convolution~\cite{figurnov2016perforatedcnns}, denoted by $\oslash$,
   which assembles only active
   pixels for computation in each pooling branch and thus avoids computing all
   pooled versions.
   }
\vspace{-4mm}
\label{fig:multiPool}
\end{figure}

In pixel-labeling tasks, the ideal spatial support for analyzing a pixel can
vary over the visual field in order to simultaneously maintain fine-grained
details and capture context.  This suggests an adaptive pooling mechanism at
pixel level, or multi-scale pooling module (\emph{MultiPool}) that chooses the
appropriate pooling size for each pixel (see e.g., \cite{kong2017recurrentscene}).
Given a collection of $P$ pooled feature maps
$\{\M_i\}_{i=1,\dots,P}$ which are computed in convolution with different dilate rates
(Fig.~\ref{fig:multiPool}),
we can generate
a MultiPool feature map ${\bf O}=\sum_i \W_i \odot\M_i$, where
$\{\W_i\}_{i=1,\dots,P}$ are spatial selection masks, and $\odot$ indicates
element-wise product between $\W_i$ and each channel of $\M_i$. We utilize the
PAG to generate binary weight $\W_i$ that selects the ``correct''
pooling region at each spatial location by
applying Eq. (\ref{eq:GumbelArgmax}).  This MultiPool module, illustrated
in Fig.~\ref{fig:multiPool}, can be inserted in place of regular pooling
with little computational overhead and learned in a latent manner using
the task-specific loss (no additional sparsity loss).

We implement pooling using a set of 3$\times$3-kernels applied at a set of
user-specified dilation rates ($[0,1,2,4,6,8,10]$)~\cite{yu2015multi}.  A dilation rate of 0 means
the input feature is simply copied into the output feature map.  In our
experiments, we observe that only a small portion of pixels are exactly copied
for the final representation without being fed into any multi-pooling branches.
Note that in Fig.~\ref{fig:multiPool}, a multiplicative gating operation is
shown for clarity, but an efficient implementation would utilize masking to
directly select pixels in a matrix multiplication implementation of the
convolutional layers in GPU or FPGA
kernel~\cite{chetlur2014cudnn,ovtcharov2015accelerating}.
Our MultiPool module is thus distinct from \cite{kong2017recurrentscene} which
use weighted averages over all intermediate feature activations from all
branches for the final feature representation. Our approach selects a single
pooling size for each pixel and hence does not require overhead of computing
all branches.

\section{Implementation and Training}
\label{sec:impl}
While our PAG unit is agnostic to network architectures, in all our experiments
we utilize ResNet~\cite{he2016deep} pre-trained on
ImageNet~\cite{deng2009imagenet} as the base model.  We
following~\cite{chen2016deeplab,kong2017recurrentscene} and increase the output
resolution of ResNet by removing the top global 7$\times$7 pooling layer and
the last two 2$\times$2 pooling layers, replacing them with atrous convolution
with dilation rate 2 and 4, respectively, to maintain a spatial sampling rate.
Such a modification thus outputs predictions at $1/8$ the input resolution.
Rather than up-sampling the output (or downsampling the ground-truth) 8$\times$
for benchmarking~\cite{chen2016deeplab,kong2017recurrentscene}, we find it
better to apply a deconvolution layer followed by two or more convolutional
layers before the final output.

We augment the training sets with random left-right flips and crops with
20-pixel margin and of size divisible by 8.  When training the model, we fix
the batch normalization, using the same constant global moments in both
training and testing.  This modification does not impact the performance and
allows a batch size of one during training (a single input image per batch).
We use the ``poly'' learning rate policy~\cite{chen2016deeplab} with a base
learning rate of $0.0002$ scaled as a function of iteration by
$(1-\frac{iter}{maxiter})^{0.9}$.
We adopt a stage-wise training strategy over all tasks, \ie training a base
model, adding PAG-based MultiPool, inserting PAG for dynamic computation progressively,
and finally decreasing $\rho$ to achieve target computational budget.
Since our goal is to explore computational parsimony in per-pixel labeling tasks,
we implement our models without ``bells and whistles'', \eg no utilization of
ensembles, no CRF as post-processing, and no external training data.
We implement our approach using the toolbox
MatConvNet~\cite{vedaldi2015matconvnet}, and train using SGD on a single Titan
X GPU\footnote{As MatConvNet itself does not provide perforated convolution, we
release the code and models implemented with multiplicative gating at
{\color{blue} \emph{
{https://github.com/aimerykong/Pixel-Attentional-Gating}}}.}.


\subsection{Per-Pixel Labeling Vision Tasks}
\label{ssec:ppl}

\noindent \textbf{Boundary Detection}
We train a base model using (binary) logistic loss.
Following~\cite{xie2015holistically,maninis2017convolutional,kong2017recurrentpixel},
we include four prediction branches at macro residual blocks (denoted by Res
$2, 3, 4, 5$) and a fusion branch for training.
To handle class imbalance, we utilize a weighted loss accumulated over the prediction
losses given by:
\begin{equation}
\small
\begin{split}
\ell_{boundary} = & -\sum_{b \in {\cal B}}   \sum_{j\in Y} \beta_{y_j}\log(P(y_j \vert \theta_b))
\end{split}
\label{eq:loss_boundary}
\end{equation}
where $b$ indexes the branches,
$\beta_+=\vert Y_-\vert/\vert Y_- \cup Y_+
\vert$, $\beta_-=1-\beta_+$;
$Y_+$ and $Y_-$ denote the set of boundary and non-boundary
annotations, respectively. $Y=Y_- \cup Y_+$ contains the indices of all pixels.
This base model is modeled after HED~\cite{xie2015holistically}
and performs similarly.

\noindent \textbf{Semantic Segmentation}
For semantic segmentation, we train a model using $K$-way cross-entropy loss as
in~\cite{chen2016deeplab,kong2017recurrentscene}:
\begin{equation}
\small
\begin{split}
\ell_{semantic} = -\sum_{i} \sum_{c=1}^{K} 1_{[y_i=c]} \cdot \log(C_i)
\end{split}
\label{eq:loss_semantic}
\end{equation}
where $C_i$ is the class prediction (from a softmax transform) at pixel $i$,
and $y_i$ is the ground-truth class label.

%
%
%

\noindent \textbf{Monocular Depth Estimation}
For monocular depth estimation,
we use combined $L_2$ and $L_1$ losses to compare the predicted
and ground-truth depth maps $\D$ and $\hat\D$ which are on a log scale:
\begin{equation}
\small
\begin{split}
\ell_{depth} = & \sum_{i=1} \Vert \D_i - \hat\D_i\Vert_2^2 + \gamma\Vert \D_i - \hat\D_i \Vert_1
\end{split}
\end{equation}
where $\gamma=2$ controls the relative importance of the two losses.  This
mixed loss penalizes large errors quadratically (the $L_2$ term) while still
assuring a non-vanishing gradient that continues to drive down small errors
(the $L_1$ term).  The idea behind our loss is similar to the reverse Huber
loss as used in \cite{laina2016deeper}, which can be understood as
concatenation of truncated ${\cal L}_2$ and ${\cal L}_1$ loss.  However, the
reverse Huber loss requires specifying a hyper-parameter for the boundary
between ${\cal L}_2$ and ${\cal L}_1$; we find our mixed loss is robust
and performs well with $\gamma \in [1,5]$.

\begin{figure}[t]
\raisebox{-\height}{\vspace{0pt}\includegraphics[width=0.23 \textwidth]{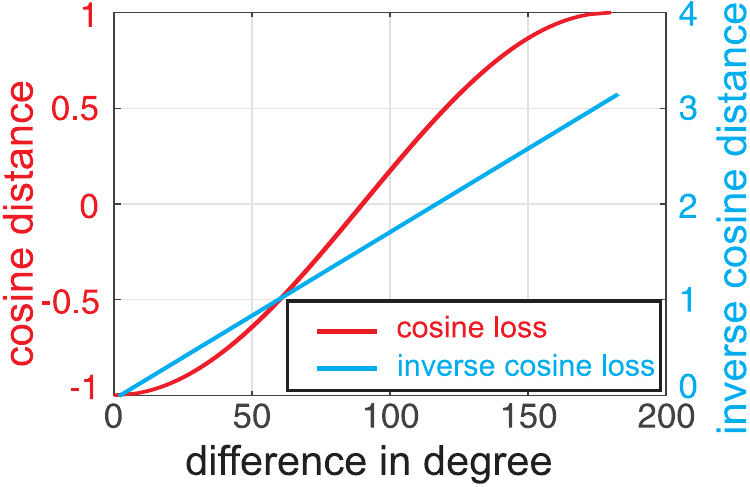}}\hfill%
\begin{minipage}[t]{0.23\textwidth}
\caption{\small
Inverse
cosine loss has a constant gradient, while the gradient of the widely used
cosine loss decreases as the prediction errors become small, preventing
further model refinement.}
\label{fig:loss_normal}
\end{minipage}
\vspace{-7mm}
\end{figure}

\noindent \textbf{Surface Normal Estimation}
To predict surface normals, we insert a final $L_2$ normalization layer so that
predicted normals have unit Euclidean length.  In the literature, cosine
distance is often used in the loss function to train the model, while
performance metrics for normal estimation measure the angular difference
between prediction $\n$ and the target normal $\hat\n$
\cite{fouhey2013data,eigen2014depth}. We
address this discrepancy by incorporating inverse cosine distance along with
cosine distance as our objective function:
\begin{equation}
\small
\begin{split}
\ell_{normal} = & \sum_i -\n_i^T\hat\n_i + \lambda \cos^{-1} (\n_i^T\hat\n_i) \\
\end{split}
\end{equation}
where $\lambda$ controls the importance of the two part and we set $\lambda=4$
throughout our experiments.  Fig.~\ref{fig:loss_normal} compares the curves of
the two losses, and we can clearly see that the inverse cosine loss always
produce meaningful gradients, whereas the popular cosine loss has ``vanishing
gradient'' issue when prediction errors become small (analogous to the mixed
$L_1/L_2$ loss for depth).

\section{Experiments}
To evaluate our method based on PAG,
we choose datasets that span a variety of per-pixel labeling tasks,
including boundary detection, semantic segmentation,
depth and surface normal estimation.
We first describe the datasets,
and then carry out experiments to determine the best architectural
configurations to exploit PAG and measure compute-performance trade-offs.
We then evaluate our best models on standard benchmarks and
show our approach achieves state-of-the-art or competitive performance
for pixel-labeling.
Finally,
we visualize the attentional maps from MultiPool and ponder maps, and
demonstrate qualitatively that our models pay more ``attention'' to specific
regions/pixels, especially on boundaries between regions, \eg semantic
segments, and regions with sharp change of depth and normal.

\subsection{Datasets}
We utilize the following benchmark datasets.

\noindent \textbf{BSDS500}~\cite{arbelaez2011contour}
is the most popular dataset for boundary detection.
It provides a standard split~\cite{arbelaez2011contour,xie2015holistically}
of 300 train-val images and 200 test images.

\noindent \textbf{NYUv2}~\cite{silberman2012indoor} consists of 1,449
  RGB-D indoor scene images of the resolution $640\times 480$ which include
  color and pixel-wise depth obtained by a Kinect sensor.  We use the
  ground-truth segmentation into 40 classes provided
  in~\cite{gupta2013perceptual} and a standard train/test split into 795 and
  654 images, respectively.
  For surface normal estimation,
  we compute the normal as target from depth using the method in
  \cite{silberman2012indoor} by fitting least-squares planes to
  neighboring sets of points in the point cloud.

\noindent \textbf{Stanford-2D-3D}~\cite{armeni2017joint} contains 1,559
  RGB panoramic images with depths, surface normal and semantic annotations
  covering six large-scale
  indoor areas from three different buildings.  We use area 3 and 4 as a
  validation set (489 panoramas) and the remaining four areas for training
  (1,070 panoramas).  The panoramas are very large (2048$\times$4096) and
  contain black void regions at top and bottom due to the spherical panoramic
  topology. We rescale them by $0.5$ and crop out the central two-thirds
  ($y \in [160, 863]$) resulting in final
  images of size 704$\times$2048-pixels.
  We randomly crop out sub-images of
  704$\times$704 resolution for training.
  Note that the surface normals in panoramic images are relative to the global coordinate
  system which cannot be determined from the image alone.
  Thus we transformed this global normal into local normal specified relative
  to the camera viewing direction (details in supplementary material).
  Note that such relative normals are also useful in scene understanding and
  reconstruction.

\noindent \textbf{Cityscapes}~\cite{cordts2016cityscapes} contains high-quality
  pixel-level annotations of images collected in street scenes from 50
  different cities. We use the standard split of training set (2,975 images) and
  validation set (500 images) for testing, respectively, labeled for 19 semantic
  classes as well as depth obtained by disparity.  The images are of high
  resolution ($1024\times2048$), and we randomly crop out sub-images of
  800$\times$800 resolution during training.

\subsection{Analysis of Pixel-wise Attentional Gating}
\label{ssec:ablation}

{
\setlength{\tabcolsep}{0.28em} 
\begin{table}[t]
\caption{ \small Ablation study for where to insert the PAG-based MultiPool module.
Experiments are from boundary detection and semantic segmentation on
BSDS500 and NYUv2 dataset, measured by $F$-score ($F_{bnd.}$) and IoU (IoU$_{seg.}$),
respectively.
Numbers are in $\%$ (higher is better).}
\centering
\vspace{-1mm}
{\scriptsize
\begin{tabular}{l | c c c c c c c c c c}
\hline
    metrics & base.
    & Res$3$ & Res$4$ & Res$5$
    & Res$6$ & Res4-5 & Res3-5
    & Res4-6 & Res5-6 \\
    \hline
$F_{bnd.}$& 79.00
 & 79.19 & \textbf{79.19} & 79.14
 & --- & 79.18 & 79.07
 & --- & --- \\
 \hline
\hline
 IoU$_{seg.}$ & 42.05
 & 44.13 & 45.67 & \textbf{46.52}
 & 45.99 & 45.48 & 44.83
 & 44.97 & 46.44  \\
\hline
\end{tabular}
}
\label{tab:ablation_NYUv2_segmentation}
\vspace{-1mm}
\end{table}
}

{
\setlength{\tabcolsep}{0.2em} 
\begin{table}[t]
\vspace{-1mm}
\caption{\small Performance comparison w.r.t computational parsimony controlled by hyper-parameter
$\rho$ on NYUv2 dataset for semantic segmentation.}
\vspace{-1mm}
\centering
{\scriptsize 
\begin{tabular}{c c c c c c c c c  c c c c c c c c} 
\hline
    \multicolumn{2}{c} {param.\&FLOPs }
    &  \multicolumn{2}{c} {truncated}
    &  \multicolumn{2}{c} {layer-skipping}
    &  \multicolumn{2}{c} {perforatedCNN}
    &  \multicolumn{2}{c} {MP@Res5 (PAG)}       \\
    \cmidrule(r){1-2}
    \cmidrule(r){3-4} \cmidrule(r){5-6} \cmidrule(r){7-8} \cmidrule(r){9-10}
$\rho$ & $\times 10^{10}$ &  IoU & acc. &  IoU & acc. & IoU & acc. &  IoU & acc.  \\
\hline
0.5 & 6.29 & 36.30 & 67.36   & 37.78 & 67.31 & 37.37 & 66.76  & 40.89 & 69.44  \\
\hline
0.7 & 8.27 & 37.69 & 67.44   & 39.84 & 69.00 & 40.09 & 68.78  & 43.61 & 71.41 \\
\hline
0.9 & 8.95 & 40.29 & 69.66   & 41.27 & 70.01 & 42.94 & 70.94  & 45.75 & 72.93 \\
\hline
1.0 & 9.63 & --- & ---   & --- & --- &       --- & --- & 46.52 & 73.50  \\
\hline
\end{tabular}
}
\label{tab:ablation_NYUv2_parsimony}
\vspace{-1mm}
\end{table}
}

We evaluate different configurations of PAG on the BSDS500 and NYUv2 datasets
for boundary detection and semantic segmentation (similar observations hold on
other tasks, see supplementary materials). The goal of these experiments is
to establish:
\begin{enumerate}[itemsep=0mm]
  \item whether our base model is comparable to state-of-the-art methods;
  \item where to insert the PAG-based MultiPool module for the best performance;
  \item how our PAG-based method for computational parsimony impacts performance,
    and how it performs compared with other related methods, \eg
    truncated ResNet and methods learning to skip/drop layers.
\end{enumerate}

\noindent{\bf Base models:}
We train our base models as described in Section~\ref{sec:impl} without PAG units.
The performance of our base model is on-par with state-of-the-art systems, achieving
IoU=$42.05\%$ on NYUv2 for semantic segmentation
(RefineNet~\cite{lin2016refinenet} achieves IoU=$44.5$ with multi-resolution input),
and $F=0.79$ on BSDS500 for boundary detection
(HED \cite{xie2015holistically} achieves $F=0.78$).
More comprehensive comparisons with other related methods are shown later in
Section \ref{ssec:comprehensiveComparison}.

\noindent{\bf MultiPool:}
Table~\ref{tab:ablation_NYUv2_segmentation} explores the effect of inserting the
MultiPool operation at different layers in the base model.  In
Table~\ref{tab:ablation_NYUv2_segmentation}, Res6 means that we insert
MultiPool module in the additional convolutional layers above the ResNet50
base.  For boundary detection, we do not initialize more convolutional layers
above the backbone, so there is no Res6. For both tasks, we observe that
including a PAG-based MultiPool module improves performance, but including more
than one MultiPool module does not offer further improvements.  We find
inserting MultiPool module at second last macro residual block (Res4 or Res5
depending on task) yields the largest gain.

For semantic segmentation, our MultiPool also outperforms the weighted
pooling in~\cite{kong2017recurrentscene}, which uses the same ResNet50
base.  We conjecture this is due to three reasons.  First, we apply the
deconvolutional layer way before the last convolutional layer for softmax input
as explained in Section~\ref{ssec:ppl}.  This increases resolution that enables
the model to see better the fine details.  Additionally, our set of pooling
regions includes finer scales (rather than using powers of 2).
Finally, the results in Table~\ref{tab:compare_seg} show that PAG with binary
masks performs slightly better (IoU=46.5 vs. IoU=46.3) than the (softmax)
weighted average operation used in \cite{kong2017recurrentscene}.

\noindent{\bf Computation-Performance Tradeoffs:} Lastly, we evaluate how our
dynamic parsimonious computation setup impacts performance and performs
compared with other baselines.  We show results of semantic segmentation on
NYUv2 dataset in Table~\ref{tab:ablation_NYUv2_parsimony}, comparing different
baselines and our models with MultiPool at macro block Res5,
\emph{MP@Res5 (PAG)} for short, which are trained with
different target computational budgets (specified by $\rho$).  The
``truncated'' baseline means we simple remove top layers of ResNet to save
computation, while ``layer-skipping'' is an implementation of
\cite{veit2017convolutional} that learns to dynamically skip a subset of
layer. ``PerforatedCNN'' is our implementation of~\cite{figurnov2016perforatedcnns}
that matches the computational budget using a learned constant gating function
(not dependent on input image).
For fair comparison, we insert MultiPool module at the top of all
the compared methods.
These results clearly suggest that the PAG approach outperforms all these
methods, demonstrating that learning dynamic computation path at the
pixel level is helpful for per-pixel labeling tasks. It is also worth noting that
PerforatedCNN does not support fully convolutional computation requiring that the
input image have a fixed size in order to learn fixed computation paths over the
image.  In contrast, our method is fully convolutional that is able to take as
input images of arbitrary size and perform computing with input-dependent
dynamic paths.

\begin{figure}[t]
\centering
   \includegraphics[width=0.95\linewidth]{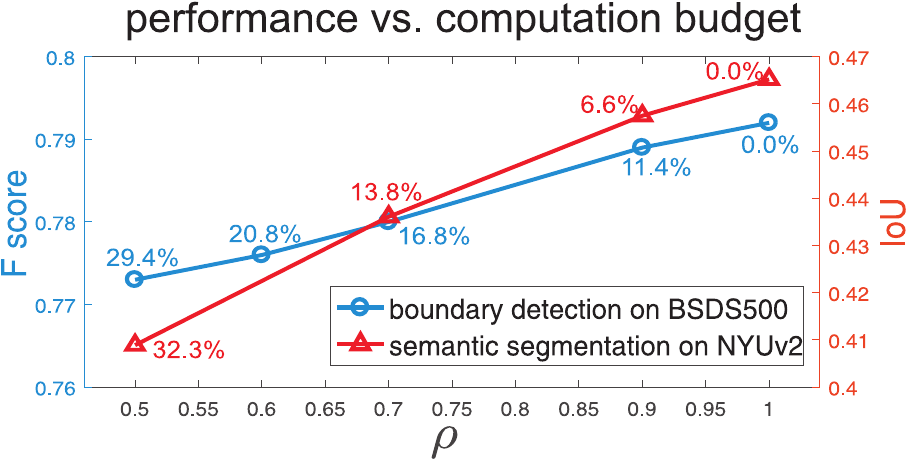}
\vspace{-4mm}
   \caption{
    \small Performance vs. computation budget (controlled by $\rho$) for
    boundary detection and semantic segmentation,
    with saved computation (\%) compared to full model as
    indicated by percentage number.
   }
\vspace{-2mm}
\label{fig:BSDS500_boundary_curves}
\end{figure}

Fig.~\ref{fig:BSDS500_boundary_curves} shows that, as we decrease the
computation budget, the performance of the PAG-based method degrades gracefully
even as the amount of computation is scaled back to 70\%, merely inducing
$2.4\%$ and $5.6\%$ performance degradation on boundary detection and semantic segmentation
compared to their full model, respectively,
i.e., $F$=$0.773$ vs. $F$=$0.792$ and $IoU$=$0.409$ vs. $IoU$=$0.465$.
Table \ref{tab:ablation_NYUv2_parsimony} highlights the comparison to
truncation and layer-skipping models adjusted to match the same computational
budget as PAG. For these approaches, performance decays much more sharply with
decreasing budget. These results also highlight that the target sparsity
parameter $\rho$ provides tight control over the actual average computation
of the model.


{
\setlength{\tabcolsep}{0.2em} 
\begin{table}[t]
\caption{ \small Comparison with the state-of-the-art methods for boundary detection
on BSDS500 dataset.
}
\centering
\vspace{-2mm}
{\scriptsize
\begin{tabular}{l | c c c c c c c c c c}
\hline
method & MP@Res4 & HED & COB & LEP & MCG & MShift & gPb-UCM & NCut & EGB \\
       &         & \cite{xie2015holistically} & \cite{maninis2017convolutional}
       & \cite{najman1996geodesic} & \cite{arbelaez2014multiscale}
       & \cite{comaniciu2002mean} & \cite{arbelaez2011contour}
       & \cite{shi2000normalized} & \cite{felzenszwalb2004efficient} \\
\hline
odsF & 0.792 & 0.780 & 0.793 & 0.757 & 0.747 & 0.601 & 0.726 & 0.641 & 0.636 \\
oisF & 0.806 & 0.796 & 0.820 & 0.793 & 0.779 & 0.644 & 0.760 & 0.674 & 0.674 \\
AP   & 0.832 & 0.834 & 0.859 & 0.828 & 0.759 & 0.493 & 0.727 & 0.447 & 0.581 \\
\hline
\end{tabular}
}
\label{tab:boundaryDetSOTA}
\vspace{-4mm}
\end{table}

\subsection{Comprehensive Benchmark Comparison}
\label{ssec:comprehensiveComparison}

{
\setlength{\tabcolsep}{0.2em} 
\begin{table}[t]
\centering
\caption{\small Semantic segmentation is measured by Intersection over Union (IoU),
pixel accuracy (acc),
and iIoU that leverages the size of segments w.r.t categories.
Results marked by $^\dagger$ are from our trained models with the released code.
}
\scriptsize
\vspace{-2mm}
\begin{tabular}{l c c c  c c c c c } 
\hline
    &  \multicolumn{2}{c} {NYUv2~\cite{silberman2012indoor}}
    &  \multicolumn{2}{c} {Stanford-2D-3D~\cite{armeni2017joint}}
    &  \multicolumn{2}{c} {Cityscapes~\cite{silberman2012indoor}}         \\
    \cmidrule(r){2-3} \cmidrule(r){4-5} \cmidrule(r){6-7}
    methods/metrics & IoU & pixel acc. & IoU & pixel acc. &  IoU & iIoU \\
\hline
baseline           & 42.1 & 71.1 &  79.5 & 92.1 & 73.8 & 54.7 \\
MP@Res5 (w-Avg.)   & 46.3 & 73.4 &  {\bf 83.7} & 93.6 & 75.8 & 56.9 \\
MP@Res5 (PAG)      & {\bf 46.5} & 73.5 &  {\bf 83.7} & {\bf 93.7} & 75.7 & 55.8 \\
\hline
MP@Res5 ($\rho$$=$0.9)& 45.8 & 72.9 &  82.8 & 93.3 & 75.0 & 55.4 \\
MP@Res5 ($\rho$$=$0.7)& 43.6 & 71.4 &  82.4 & 93.2 & 72.6 & 55.1 \\
MP@Res5 ($\rho$$=$0.5)\ \ \ \ \ \ \ \ \ \
                      & 40.9 & 69.4 &  81.8 & 92.9 & 70.8 & 53.2 \\
\hline
PerspectiveParsing~\cite{kong2017recurrentscene} & 44.5 & 72.1 &  76.5 & 91.0 & 75.4 & 56.8 \\
DeepLab~\cite{chen2016deeplab} & --- & --- &  69.8$^\dagger$& 88.0$^\dagger$ & 71.4 & 51.6 \\
LRR~\cite{ghiasi2016laplacian} & --- & --- &  --- & --- & 70.0 & 48.0 \\
PSPNet~\cite{zhao2016pyramid}  & --- & --- & 67.4$^\dagger$ & 87.6$^\dagger$ & {\bf 78.7} & {\bf 60.4} \\
RefineNet-Res50~\cite{lin2016refinenet}& 43.8 & --- & --- & --- & ---&---\\
RefineNet-Res152~\cite{lin2016refinenet}& {\bf 46.5} & {\bf 73.6} & --- & --- & --- &---\\
\hline
\end{tabular}
\label{tab:compare_seg}
\vspace{-2mm}
\end{table}
}

{
\setlength{\tabcolsep}{0.3em} 
\begin{table}[t]
\centering
\caption{\footnotesize Depth estimation is measured by standard threshold accuracy,
\ie the percentage ($\%$) of predicted pixel depths $d_i$ s.t.
$\delta=\max(\frac{d_i}{\hat d_i},\frac{\hat d_i}{d_i})<\tau$,
where $\tau=\{1.25, 1.25^2, 1.25^3\}$.
Methods with $^{*}$ use $\sim$100k extra images to train.
}
\vspace{-2mm}
\tiny
\begin{tabular}{l  c c c c c c c c c  c c c c} 
\hline
    & \multicolumn{3}{c} {NYUv2~\cite{silberman2012indoor}}
    &  \multicolumn{3}{c} {Stanford-2D-3D~\cite{armeni2017joint}}
    &  \multicolumn{3}{c} {Cityscapes~\cite{silberman2012indoor}}       \\
    \cmidrule(r){2-4}     \cmidrule(r){5-7} \cmidrule(r){8-10}
    methods/metric  & $1.25$ & $1.25^2$ &$1.25^3$
    & $1.25$ & $1.25^2$ &$1.25^3$
    & $1.25$ & $1.25^2$ &$1.25^3$  \\
\hline
baseline              & 71.1 & 93.2 & 98.5 & 73.1 & 92.1 & 97.5 & 29.0 & 53.8 & 75.8 \\
MP@Res5 (w-Avg.)      & 74.5 & 94.4 & {\bf 98.8} & 77.5 & {\bf 94.1} & 97.9 & 33.7 & 65.9 & 76.9 \\
MP@Res5 (PAG)         & 75.1 & 94.4 & {\bf 98.8} & {\bf 77.6} & {\bf 94.1}  & {\bf 97.9} & {\bf 34.6}
& {\bf 66.2} & {\bf 77.2} \\
\hline
MultiPool ($\rho$$=$0.9)& 74.5 & 94.4 & 98.8 & 77.3 & 93.9 & 97.8 & 34.5 & 65.7 & 76.9 \\
MultiPool ($\rho$$=$0.7)& 71.0 & 93.3 & 98.5 & 75.4 & 92.8 & 97.6 & 32.0 & 63.5 & 75.8 \\
MultiPool ($\rho$$=$0.5)& 67.3 & 91.0 & 97.7 & 72.7 & 91.3 & 97.1 & 28.7 & 58.7 & 71.6 \\
\hline
Liu~\cite{laina2016deeper}
    & 61.4 & 88.3 & 97.1 & --- & ---  & --- & --- & --- & --- \\
Ladicky$^{*}$~\cite{ladicky2014pulling}
    & 54.2 & 82.9 & 94.0 & --- & ---  & --- & --- & --- & --- \\
Eigen$^{*}$~\cite{eigen2014depth}
    & 61.4 & 88.8 & 97.2 & --- & ---  & --- & --- & --- & --- \\
Eigen$^{*}$~\cite{eigen2015predicting}
    & 76.9 & 95.0 & 98.8 & --- & ---  & --- & --- & --- & --- \\
Laina$^{*}$~\cite{laina2016deeper}
    & {\bf 81.1} & {\bf 95.3} & {\bf 98.8} & --- & ---  & --- & --- & --- & --- \\
\hline
\end{tabular}
\label{tab:compare_depth}
\vspace{-1mm}
\end{table}
}

We now compare our models under different degrees of computational parsimony
($\rho$=$\{0.5, 0.7, 0.9, 1.0\}$) with other state-of-the-art systems for pixel labeling.

Taking boundary detection as the first task,
we quantitatively compare our model to
COB~\cite{maninis2017convolutional},
HED~\cite{xie2015holistically},
gPb-UCM~\cite{arbelaez2011contour},
LEP~\cite{najman1996geodesic},
UCM~\cite{arbelaez2011contour},
NCuts~\cite{shi2000normalized},
EGB~\cite{felzenszwalb2004efficient},
MCG~\cite{arbelaez2014multiscale}
and the mean shift (MShift) algorithm~\cite{comaniciu2002mean}.
Table~\ref{tab:boundaryDetSOTA} shows comparison to
all the methods (PR curves in supplementary material),
demonstrating our model achieves state-of-the-art performance.
Note that our model has the same backbone architecture of HED~\cite{xie2015holistically},
but outperforms it with our MultiPool module which increases receptive fields
at higher levels.
Our model performs on par with COB~\cite{maninis2017convolutional},
which uses auxiliary losses for oriented boundary detection.
Note that it is possible to surpass human performance with
sophisticated techniques~\cite{kokkinos2015pushing}, but we don't pursue
this as it is out the scope of this paper.

Table~\ref{tab:compare_seg}, \ref{tab:compare_depth} and \ref{tab:compare_normal}
show the comprehensive comparisons on the tasks of semantic segmentation,
monocular depth and surface normal estimation, respectively.
In addition to comparing with state-of-the-art methods, we also show the result
of MultiPool module with softmax weighted average operation, termed by \emph{MP@Res5 (w-Avg.)}.
Interestingly, MultiPool performs slightly better when equipped with PAG than
the weighted average fusion.  We attribute this to the facts that, longer
training has been done in the stage-wise training strategy, and PAG unit also
constrains the information flow to train specific branches.

Our baseline model achieves performance on par with recent methods for all tasks.
When inserting the MultiPool module, we improve even further and surpass the
compared methods for most tasks and datasets.  In particular, on datasets with
large perspective images, \ie Stanford-2D-3D and Cityscapes, the MultiPool
module shows greater improvement. Reducing the computation 20-30\% only
yields a performance drop of 3-5\% generally.

For depth and surface normal estimation tasks, our baseline models also perform
very well. This is notable since we don't leverage multi-task learning
(unlike Eigen~\cite{eigen2015predicting}) and do not use extra images to
augment training set (unlike most methods for depth estimation using $\sim$100k extra
images to augment the training set as shown in Table~\ref{tab:compare_depth}).
We attribute this to the combination of
the proposed PAG MultiPool and carefully designed losses for depth and
surface normal estimation.

{
\setlength{\tabcolsep}{0.2em} 
\begin{table}[t]
\centering
\caption{\footnotesize Surface normal estimation is measured
by mean angular error and
the percentage of prediction error within $t^{\circ}$ degree
where $t=\{11.25, 22.50, 30.00\}$.
Smaller ang. err. means better performance as marked by $\downarrow$.
}
\vspace{-2mm}
\tiny
\begin{tabular}{l c c c c c c c c c c c c c} 
\hline
    & \multicolumn{4}{c} {NYUv2~\cite{silberman2012indoor}}
    &  \multicolumn{4}{c} {Stanford-2D-3D~\cite{armeni2017joint}}     \\
    \cmidrule(r){2-5}     \cmidrule(r){6-9}
    methods/metrics & ang. err.$\downarrow$ & $11.25^{\circ}$ & $22.50^{\circ}$ & $30.00^{\circ}$
    & ang. err.$\downarrow$ & $11.25^{\circ}$ & $22.50^{\circ}$ & $30.00^{\circ}$ \\
\hline
baseline        & 22.3 & 34.4 & 62.5 & 74.4 & 19.0 & 51.5 & 68.6 & 76.3 \\
MP@Res5 (w-Avg.)& 21.9 & 35.9 & 63.8 & 75.3 & {\bf 16.5} & 58.2 & {\bf 74.2} & {\bf 80.4} \\
MP@Res5 (PAG)   & {\bf 21.7} & 36.1 & {\bf 64.2} & {\bf 75.5} & {\bf 16.5} & {\bf 58.3}
& {\bf 74.2} & {\bf 80.4} \\
\hline
MP@Res5 ($\rho$$=$0.9)& 21.9 & 35.9 & 63.9 & 75.4 & 16.7 & 57.5 & 73.7 & 80.1 \\
MP@Res5 ($\rho$$=$0.7)& 22.5 & 34.7 & 62.5 & 74.1 & 17.0 & 56.5 & 73.1 & 79.7 \\
MP@Res5 ($\rho$$=$0.5)& 23.6 & 31.9 & 59.7 & 71.8 & 17.7 & 54.7 & 71.4 & 78.5  \\
\hline
Fouhey~\cite{fouhey2013data}
                      & 35.3 & 16.4 & 36.6 & 48.2 & ---  & --- & --- & --- \\
Ladicky~\cite{Ladicky2014discriminatively}
                      & 35.5 & 24.0 & 45.6 & 55.9 & ---  & --- & --- & --- \\
Wang~\cite{wang2015designing}
                      & 28.8 & 35.2 & 57.1 & 65.5 & ---  & --- & --- & --- \\
Eigen~\cite{eigen2015predicting}
                      & 22.2 & {\bf 38.6} & 64.0 & 73.9 & ---  & --- & --- & --- \\
\hline
\end{tabular}
\hspace{0.5mm}
\label{tab:compare_normal}
\vspace{-1mm}
\end{table}
}

\subsection{Qualitative Visualization}
We visualize the prediction and attention maps in Fig. \ref{fig:splashFigure}
for the four datasets, respectively. We find that the binary attention maps
are qualitatively similar across layers and hence summarize them with a
``ponder map'' by summing maps across layers (per-layer maps can be found in
the supplementary material).  We can see our models allocate more computation
on the regions/pixels which are likely sharp transitions, \eg boundaries
between semantic segments, depth discontinuties and normal discontinuities
(e.g. between wall and ceiling).

\section{Conclusion and Future Work}
In this paper, we have studied the problem of dynamic inference for pixel
labeling tasks under limited computation budget with a deep CNN network.  To
achieve this, we propose a Pixel-wise Attentional Gating unit (PAG) that learns
to generate sparse binary masks that control computation at each layer on a
per-pixel basis. Our approach differs from previous methods in demonstrating
improved performance on pixel labeling tasks using spatially varying computation
trained with simple task-specific loss. This makes our approach a good candidate
for general use as it is agnostic to tasks and architectures, and avoids more
complicated reinforcement learning-style approaches, instead relying on a simple,
easy-to-set sparsity target that correlates closely with empirical computational
cost.  As our PAG is based on a generic attention mechanism, we anticipate
future work might explore task-driven constraints for further improvements and
savings.

{\small \noindent{\bf Acknowledgement} This project is supported by NSF grants
IIS-1618806, IIS-1253538, a hardware donation from NVIDIA and
Google Graduate Research Award.
}

{\small
\bibliographystyle{ieee}
\bibliography{egbib}

\begin{thebibliography}{10}\itemsep=-1pt

\bibitem{arbelaez2011contour}
P.~Arbelaez, M.~Maire, C.~Fowlkes, and J.~Malik.
\newblock Contour detection and hierarchical image segmentation.
\newblock {\em IEEE Transactions on Pattern Analysis and Machine Intelligence
  (PAMI)}, 33(5):898--916, 2011.

\bibitem{arbelaez2014multiscale}
P.~Arbel{\'a}ez, J.~Pont-Tuset, J.~T. Barron, F.~Marques, and J.~Malik.
\newblock Multiscale combinatorial grouping.
\newblock In {\em Proceedings of the IEEE Conference on Computer Vision and
  Pattern Recognition (CVPR)}, pages 328--335, 2014.

\bibitem{armeni2017joint}
I.~Armeni, S.~Sax, A.~R. Zamir, and S.~Savarese.
\newblock Joint 2d-3d-semantic data for indoor scene understanding.
\newblock In {\em Proceedings of the IEEE International Conference on Computer
  Vision (ICCV)}, 2017.

\bibitem{bansal2016marr}
A.~Bansal, B.~Russell, and A.~Gupta.
\newblock Marr revisited: 2d-3d alignment via surface normal prediction.
\newblock In {\em Proceedings of the IEEE Conference on Computer Vision and
  Pattern Recognition (CVPR)}, pages 5965--5974, 2016.

\bibitem{carreira2017model}
M.~A. Carreira-Perpin{\'a}n and Y.~Idelbayev.
\newblock Model compression as constrained optimization, with application to
  neural nets.
\newblock {\em Part III: Pruning. arXiv}, 2017.

\bibitem{chen2016deeplab}
L.-C. Chen, G.~Papandreou, I.~Kokkinos, K.~Murphy, and A.~L. Yuille.
\newblock Deeplab: Semantic image segmentation with deep convolutional nets,
  atrous convolution, and fully connected crfs.
\newblock {\em IEEE Transactions on Pattern Analysis and Machine Intelligence
  (PAMI)}, 2016.

\bibitem{chen2016attention}
L.-C. Chen, Y.~Yang, J.~Wang, W.~Xu, and A.~L. Yuille.
\newblock Attention to scale: Scale-aware semantic image segmentation.
\newblock In {\em Proceedings of the IEEE conference on Computer Vision and
  Pattern Recognition (CVPR)}, pages 3640--3649, 2016.

\bibitem{chetlur2014cudnn}
S.~Chetlur, C.~Woolley, P.~Vandermersch, J.~Cohen, J.~Tran, B.~Catanzaro, and
  E.~Shelhamer.
\newblock cudnn: Efficient primitives for deep learning.
\newblock {\em arXiv preprint arXiv:1410.0759}, 2014.

\bibitem{comaniciu2002mean}
D.~Comaniciu and P.~Meer.
\newblock Mean shift: A robust approach toward feature space analysis.
\newblock {\em IEEE Transactions on Pattern Analysis and Machine Intelligence
  (PAMI)}, 24(5):603--619, 2002.

\bibitem{cordts2016cityscapes}
M.~Cordts, M.~Omran, S.~Ramos, T.~Rehfeld, M.~Enzweiler, R.~Benenson,
  U.~Franke, S.~Roth, and B.~Schiele.
\newblock The cityscapes dataset for semantic urban scene understanding.
\newblock In {\em Proceedings of the IEEE Conference on Computer Vision and
  Pattern Recognition (CVPR)}, pages 3213--3223, 2016.

\bibitem{deng2009imagenet}
J.~Deng, W.~Dong, R.~Socher, L.-J. Li, K.~Li, and L.~Fei-Fei.
\newblock Imagenet: A large-scale hierarchical image database.
\newblock In {\em Proceedings of the IEEE Conference on Computer Vision and
  Pattern Recognition (CVPR)}, pages 248--255. IEEE, 2009.

\bibitem{eigen2015predicting}
D.~Eigen and R.~Fergus.
\newblock Predicting depth, surface normals and semantic labels with a common
  multi-scale convolutional architecture.
\newblock In {\em Proceedings of the IEEE International Conference on Computer
  Vision (ICCV)}, pages 2650--2658, 2015.

\bibitem{eigen2014depth}
D.~Eigen, C.~Puhrsch, and R.~Fergus.
\newblock Depth map prediction from a single image using a multi-scale deep
  network.
\newblock In {\em Advances in Neural Information Processing Systems (NIPS)},
  2014.

\bibitem{felzenszwalb2004efficient}
P.~F. Felzenszwalb and D.~P. Huttenlocher.
\newblock Efficient graph-based image segmentation.
\newblock {\em International Journal of Computer Vision (IJCV)},
  59(2):167--181, 2004.

\bibitem{figurnov2017spatially}
M.~Figurnov, M.~D. Collins, Y.~Zhu, L.~Zhang, J.~Huang, D.~Vetrov, and
  R.~Salakhutdinov.
\newblock Spatially adaptive computation time for residual networks.
\newblock In {\em Proceedings of the IEEE Conference on Computer Vision and
  Pattern Recognition (CVPR)}, 2017.

\bibitem{figurnov2016perforatedcnns}
M.~Figurnov, A.~Ibraimova, D.~P. Vetrov, and P.~Kohli.
\newblock Perforatedcnns: Acceleration through elimination of redundant
  convolutions.
\newblock In {\em Advances in Neural Information Processing Systems (NIPS)},
  pages 947--955, 2016.

\bibitem{fouhey2013data}
D.~F. Fouhey, A.~Gupta, and M.~Hebert.
\newblock Data-driven 3d primitives for single image understanding.
\newblock In {\em IEEE International Conference on Computer Vision (ICCV)},
  pages 3392--3399. IEEE, 2013.

\bibitem{ghiasi2016laplacian}
G.~Ghiasi and C.~C. Fowlkes.
\newblock Laplacian pyramid reconstruction and refinement for semantic
  segmentation.
\newblock In {\em European Conference on Computer Vision (ECCV)}, 2016.

\bibitem{gumbel2012statistics}
E.~J. Gumbel.
\newblock {\em Statistics of extremes}.
\newblock Courier Corporation, 2012.

\bibitem{gupta2013perceptual}
S.~Gupta, P.~Arbelaez, and J.~Malik.
\newblock Perceptual organization and recognition of indoor scenes from rgb-d
  images.
\newblock In {\em Proceedings of the IEEE Conference on Computer Vision and
  Pattern Recognition (CVPR)}, 2013.

\bibitem{han2015deep}
S.~Han, H.~Mao, and W.~J. Dally.
\newblock Deep compression: Compressing deep neural networks with pruning,
  trained quantization and huffman coding.
\newblock In {\em International Conference on Learning Representations (ICLR)},
  2016.

\bibitem{he2016deep}
K.~He, X.~Zhang, S.~Ren, and J.~Sun.
\newblock Deep residual learning for image recognition.
\newblock In {\em Proceedings of the IEEE Conference on Computer Vision and
  Pattern Recognition (CVPR)}, pages 770--778, 2016.

\bibitem{hinton2015distilling}
G.~Hinton, O.~Vinyals, and J.~Dean.
\newblock Distilling the knowledge in a neural network.
\newblock In {\em NIPS Workship in Deep Learning and Representation Learning},
  2015.

\bibitem{huang2017densely}
G.~Huang, Z.~Liu, K.~Q. Weinberger, and L.~van~der Maaten.
\newblock Densely connected convolutional networks.
\newblock In {\em Proceedings of the IEEE conference on Computer Vision and
  Pattern Recognition (CVPR)}, 2017.

\bibitem{huang2016deep}
G.~Huang, Y.~Sun, Z.~Liu, D.~Sedra, and K.~Q. Weinberger.
\newblock Deep networks with stochastic depth.
\newblock In {\em European Conference on Computer Vision (ECCV)}, pages
  646--661. Springer, 2016.

\bibitem{iandola2016squeezenet}
F.~N. Iandola, S.~Han, M.~W. Moskewicz, K.~Ashraf, W.~J. Dally, and K.~Keutzer.
\newblock Squeezenet: Alexnet-level accuracy with 50x fewer parameters and
  $<$0.5 mb model size.
\newblock In {\em International Conference on Learning Representations (ICLR)},
  2017.

\bibitem{ioffe2015batch}
S.~Ioffe and C.~Szegedy.
\newblock Batch normalization: Accelerating deep network training by reducing
  internal covariate shift.
\newblock In {\em International Conference on Machine Learning (ICML)}, pages
  448--456, 2015.

\bibitem{jang2016categorical}
E.~Jang, S.~Gu, and B.~Poole.
\newblock Categorical reparameterization with gumbel-softmax.
\newblock In {\em International Conference on Learning Representations (ICLR)},
  2017.

\bibitem{kokkinos2015pushing}
I.~Kokkinos.
\newblock Pushing the boundaries of boundary detection using deep learning.
\newblock In {\em International Conference on Learning Representations (ICLR)},
  2016.

\bibitem{kong2017low}
S.~Kong and C.~Fowlkes.
\newblock Low-rank bilinear pooling for fine-grained classification.
\newblock In {\em 2017 IEEE Conference on Computer Vision and Pattern
  Recognition (CVPR)}, pages 7025--7034. IEEE, 2017.

\bibitem{kong2017recurrentpixel}
S.~Kong and C.~Fowlkes.
\newblock Recurrent pixel embedding for instance grouping.
\newblock In {\em Proceedings of the IEEE conference on Computer Vision and
  Pattern Recognition (CVPR)}, 2017.

\bibitem{kong2017recurrentscene}
S.~Kong and C.~Fowlkes.
\newblock Recurrent scene parsing with perspective understanding in the loop.
\newblock In {\em Proceedings of the IEEE conference on Computer Vision and
  Pattern Recognition (CVPR)}, 2018.

\bibitem{ladicky2014pulling}
L.~Ladicky, J.~Shi, and M.~Pollefeys.
\newblock Pulling things out of perspective.
\newblock In {\em Proceedings of the IEEE Conference on Computer Vision and
  Pattern Recognition (CVPR)}, 2014.

\bibitem{Ladicky2014discriminatively}
L.~Ladicky, B.~Zeisl, and M.~Pollefeys.
\newblock Discriminatively trained dense surface normal estimation.
\newblock In {\em European Conference on Computer Vision (ECCV)}, pages
  468--484. Springer, 2014.

\bibitem{laina2016deeper}
I.~Laina, C.~Rupprecht, V.~Belagiannis, F.~Tombari, and N.~Navab.
\newblock Deeper depth prediction with fully convolutional residual networks.
\newblock In {\em International Conference on 3D Vision (3DV)}, pages 239--248.
  IEEE, 2016.

\bibitem{li2015depth}
B.~Li, C.~Shen, Y.~Dai, A.~van~den Hengel, and M.~He.
\newblock Depth and surface normal estimation from monocular images using
  regression on deep features and hierarchical crfs.
\newblock In {\em Proceedings of the IEEE Conference on Computer Vision and
  Pattern Recognition (CVPR)}, pages 1119--1127, 2015.

\bibitem{lin2016refinenet}
G.~Lin, A.~Milan, C.~Shen, and I.~Reid.
\newblock Refinenet: Multi-path refinement networks with identity mappings for
  high-resolution semantic segmentation.
\newblock In {\em Proceedings of the IEEE Conference on Computer Vision and
  Pattern Recognition (CVPR)}, 2017.

\bibitem{liu2015deep}
F.~Liu, C.~Shen, and G.~Lin.
\newblock Deep convolutional neural fields for depth estimation from a single
  image.
\newblock In {\em Proceedings of the IEEE Conference on Computer Vision and
  Pattern Recognition (CVPR)}, pages 5162--5170, 2015.

\bibitem{maddison2016concrete}
C.~J. Maddison, A.~Mnih, and Y.~W. Teh.
\newblock The concrete distribution: A continuous relaxation of discrete random
  variables.
\newblock In {\em International Conference on Learning Representations (ICLR)},
  2017.

\bibitem{mallya2017packnet}
A.~Mallya and S.~Lazebnik.
\newblock Packnet: Adding multiple tasks to a single network by iterative
  pruning.
\newblock In {\em Proceedings of the IEEE Conference on Computer Vision and
  Pattern Recognition (CVPR)}, 2018.

\bibitem{maninis2017convolutional}
K.-K. Maninis, J.~Pont-Tuset, P.~Arbelaez, and L.~Van~Gool.
\newblock Convolutional oriented boundaries: From image segmentation to
  high-level tasks.
\newblock {\em IEEE Transactions on Pattern Analysis and Machine Intelligence
  (PAMI)}, 2017.

\bibitem{molchanov2016pruning}
P.~Molchanov, S.~Tyree, T.~Karras, T.~Aila, and J.~Kautz.
\newblock Pruning convolutional neural networks for resource efficient
  inference.
\newblock In {\em International Conference on Learning Representations (ICLR)},
  2017.

\bibitem{nair2010rectified}
V.~Nair and G.~E. Hinton.
\newblock Rectified linear units improve restricted boltzmann machines.
\newblock In {\em International Conference on Machine Learning (ICML)}, pages
  807--814, 2010.

\bibitem{najman1996geodesic}
L.~Najman and M.~Schmitt.
\newblock Geodesic saliency of watershed contours and hierarchical
  segmentation.
\newblock {\em IEEE Transactions on Pattern Analysis and Machine Intelligence
  (PAMI)}, 18(12):1163--1173, 1996.

\bibitem{ovtcharov2015accelerating}
K.~Ovtcharov, O.~Ruwase, J.-Y. Kim, J.~Fowers, K.~Strauss, and E.~S. Chung.
\newblock Accelerating deep convolutional neural networks using specialized
  hardware.
\newblock {\em Microsoft Research Whitepaper}, 2(11), 2015.

\bibitem{shazeer2017outrageously}
N.~Shazeer, A.~Mirhoseini, K.~Maziarz, A.~Davis, Q.~Le, G.~Hinton, and J.~Dean.
\newblock Outrageously large neural networks: The sparsely-gated
  mixture-of-experts layer.
\newblock In {\em International Conference on Learning Representations (ICLR)},
  2017.

\bibitem{shi2000normalized}
J.~Shi and J.~Malik.
\newblock Normalized cuts and image segmentation.
\newblock {\em IEEE Transactions on Pattern Analysis and Machine Intelligence
  (PAMI)}, 22(8):888--905, 2000.

\bibitem{silberman2012indoor}
N.~Silberman, D.~Hoiem, P.~Kohli, and R.~Fergus.
\newblock Indoor segmentation and support inference from rgbd images.
\newblock In {\em European Conference on Computer Vision (ECCV)}, pages
  746--760. Springer, 2012.

\bibitem{simonyan2014very}
K.~Simonyan and A.~Zisserman.
\newblock Very deep convolutional networks for large-scale image recognition.
\newblock In {\em International Conference on Learning Representations (ICLR)},
  2015.

\bibitem{vedaldi2015matconvnet}
A.~Vedaldi and K.~Lenc.
\newblock Matconvnet: Convolutional neural networks for matlab.
\newblock In {\em Proceedings of the 23rd ACM International Conference on
  Multimedia}, pages 689--692. ACM, 2015.

\bibitem{veit2017convolutional}
A.~Veit and S.~Belongie.
\newblock Convolutional networks with adaptive computation graphs.
\newblock In {\em European Conference on Computer Vision (ECCV)}, 2018.

\bibitem{veit2016residual}
A.~Veit, M.~J. Wilber, and S.~Belongie.
\newblock Residual networks behave like ensembles of relatively shallow
  networks.
\newblock In {\em Advances in Neural Information Processing Systems (NIPS)},
  pages 550--558, 2016.

\bibitem{wang2015designing}
X.~Wang, D.~Fouhey, and A.~Gupta.
\newblock Designing deep networks for surface normal estimation.
\newblock In {\em Proceedings of the IEEE Conference on Computer Vision and
  Pattern Recognition (CVPR)}, pages 539--547, 2015.

\bibitem{wang2017skipnet}
X.~Wang, F.~Yu, Z.-Y. Dou, and J.~E. Gonzalez.
\newblock Skipnet: Learning dynamic routing in convolutional networks.
\newblock In {\em European Conference on Computer Vision (ECCV)}, 2018.

\bibitem{wu2017blockdrop}
Z.~Wu, T.~Nagarajan, A.~Kumar, S.~Rennie, L.~S. Davis, K.~Grauman, and
  R.~Feris.
\newblock Blockdrop: Dynamic inference paths in residual networks.
\newblock In {\em Proceedings of the IEEE Conference on Computer Vision and
  Pattern Recognition (CVPR)}, 2017.

\bibitem{xie2015holistically}
S.~Xie and Z.~Tu.
\newblock Holistically-nested edge detection.
\newblock In {\em Proceedings of the IEEE International Conference on Computer
  Vision (ICCV)}, pages 1395--1403, 2015.

\bibitem{yu2015multi}
F.~Yu and V.~Koltun.
\newblock Multi-scale context aggregation by dilated convolutions.
\newblock In {\em International Conference on Learning Representations (ICLR)},
  2016.

\bibitem{zhao2016pyramid}
H.~Zhao, J.~Shi, X.~Qi, X.~Wang, and J.~Jia.
\newblock Pyramid scene parsing network.
\newblock In {\em Proceedings of the IEEE Conference on Computer Vision and
  Pattern Recognition (CVPR)}, 2017.

\end{thebibliography}
}

\clearpage\begin{center}
 {\large \textbf{Pixel-wise Attentional Gating for Scene Parsing} \\
 \emph{(Appendix)} }
\end{center}

\setcounter{section}{0}

\begin{abstract}
In the supplementary material,
we first present in detail how to transform the (unpredictable) global surface
normals into (predictable) local normals in panoramic images.
We then show more ablation studies on the loss functions introduced in the main paper and
MultiPool module with/without PAG unit.
Finally,
we provide more qualitative visualization of the results for various pixel-labeling tasks,
as well as the attentional ponder maps and MultiPool maps.
\end{abstract}

\section{Local Surface Normal in Panoramas}

\begin{figure*}[h]
\centering
   \includegraphics[width=0.99\linewidth]{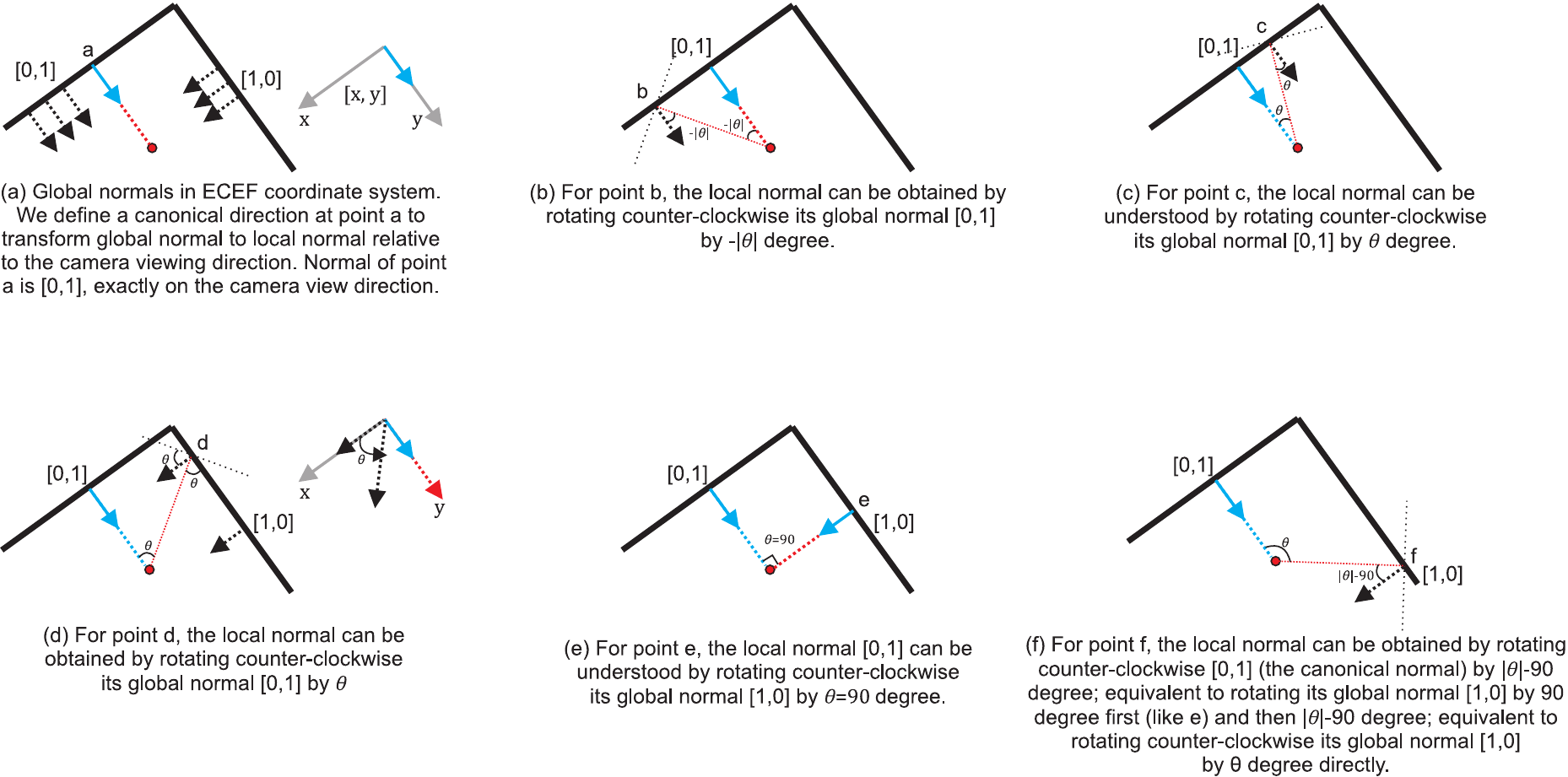}
   \caption{An illustration of how to transform global normals into local normals
   specified relative to the camera viewing direction.
   Global normals in Stanford-2D-3D dataset~\cite{armeni2017joint} are in Earth-Centered-Earth-Fixed (ECEF) coordinate system.
   }
\label{fig:suppl_illustrate_normal_transform_cases}
\end{figure*}

\begin{figure*}[h]
\centering
   \includegraphics[width=0.99\linewidth]{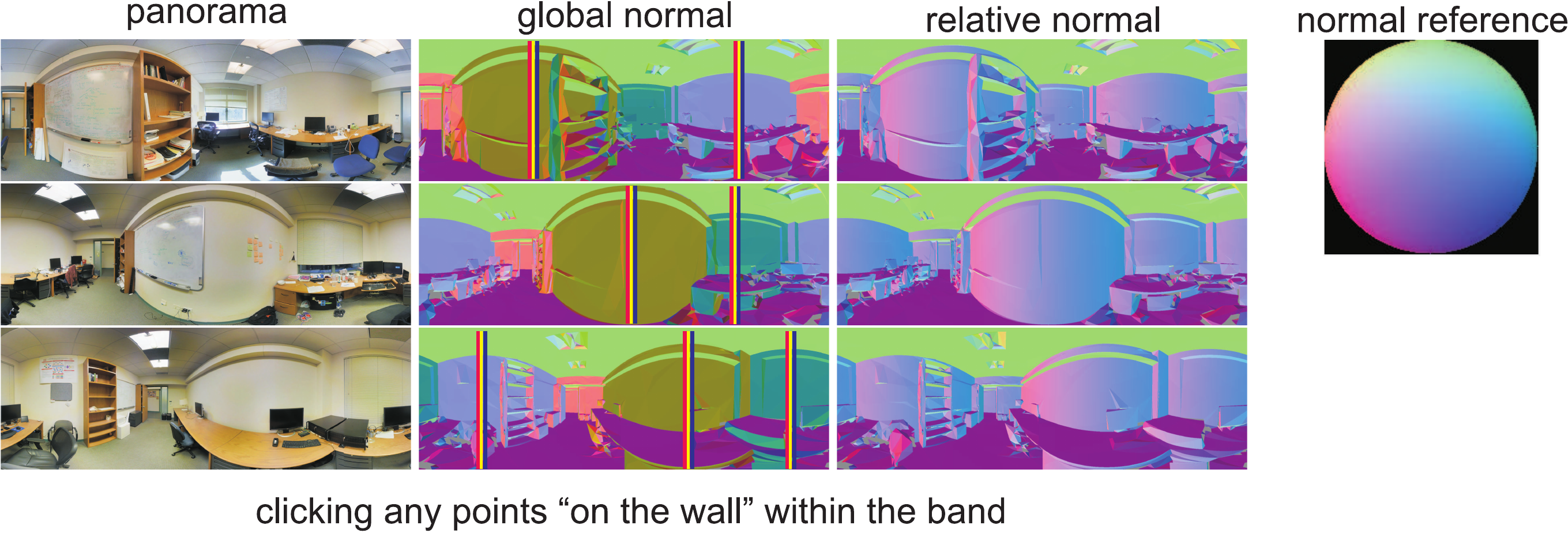}
   \caption{We draw the color bands on global normal maps to indicate points
   of the wall within these bands can be treated as canonical points,
   whose normals exactly face towards the camera.
   For a human annotator, these points can be easily detected by looking at the shape of the room.
   Our interface allows an annotator to click a point supposedly within any one of the bands,
   and through the coordinate transform,
   such global normals can be transformed into predictable, relative normals.
   The rightmost pie chart provides reference on the local surface normal relative to
   the camera viewing direction.
   }
\label{fig:suppl_illustrate_pano_normal}
\end{figure*}

Stanford-2D-3D~\cite{armeni2017joint} provides cylindrical panoramas with global surface normals,
which are in a global Earth-Centered-Earth-Fixed coordinate system (ECEF). For example,
the normals for a wall have the same direction pointing to the true north.
However,
such global coordinate system is impossible to determine from a single image,
and thus the global normals are unpredictable purely based on panoramic image data alone.
For this reason,
we propose to transform the global normals into ``local normals'' which are relative
to the camera viewing direction.
We note that such a predictability makes relative normals more useful in scene
understanding and reconstruction.

For a cylindrical panorama, we assume the vertical axis of the panorama is
aligned with the global coordinate frame.  Given a global normal at a pixel $\n = [x,y,z]^T$,
we can apply a rotation matrix $\R$ in the horizontal plane ($x$ and $y$) to
obtain its local normal $[x',y',z]^T$ in the ``camera viewing'' coordinate
system:
\begin{equation}
\begin{split}
\begin{bmatrix} x' \\ y' \\ \midrule z \\ \end{bmatrix}
=
\begin{bmatrix} \R, \0 \\ \0, 1 \\ \end{bmatrix}
\begin{bmatrix} x \\ y \\ \midrule z \\ \end{bmatrix}
\end{split}
\end{equation}
where $z$ is the variable for vertical direction.

We would like to determine the appropriate rotation matrices for all pixels
where each pixel has its own rotation matrix which is controlled by a single
signed angle parameter $\theta$.  For a cylindrical panorama, the relative
difference in viewing direction between two image locations is completely
specified by their horizontal separation in image coordinates.
Therefore, to determine the set of rotations, we simply need to specify an
origin for which the rotation is $\theta=0$, i.e., a canonical point whose
surface normal points exactly to the camera.
Given the rotation matrix for the canonical point, the rotation angle for
remaining points can be calculated as $2\pi \frac{\triangle W}{W}$, where $W$
is the width of panorama and $\triangle W$ is the offset from the target pixel
to the canonical pixel (with sign).
Fig.~\ref{fig:suppl_illustrate_normal_transform_cases} illustrates the
principle behind our methodology.

We note that it is straightforward to identify canonical points manually by
choosing a flat vertical surface (e.g., a wall) and selecting the point on
it which is nearest to the camera (e.g. shape of panoramic topology).
An automated method can be built with the same rule based on semantic
annotation and depth map.
However,
the automated method may suffer from cluttered scene (e.g. boards and bookcase on
the wall),
yet such manual annotation enables us to visualize what the local normals would look
like if we clicked the ``wrong'' points\footnote{Clicking on
the ``wrong'' points will leads to some normals pointing outwards the camera.}.
From the three random panoramic images as shown in
Fig.~\ref{fig:suppl_illustrate_pano_normal}, we can see such canonical points
lie in the color bands (manually drawn for demonstration) noted in the figure.
They are easy to detect by eye based on the warping effect due to
panoramic topology.  We made an easy-to-use tool to click a canonical point for
each panoramic image.  Fig.~\ref{fig:suppl_illustrate_pano_normal} demonstrates
the resulting transformed normals after an annotator has clicked on some point
in the color band.
We note that annotating each panoramic image costs less than 5 seconds,
and it required less than three hours to carefully annotate all 1,559
panoramas in the dataset.
We also compare the annotation when clicking on different canonical points (at different
color bands) for the same image,
and the maximum difference of normals for all spatial locations is only less than $8^{\circ}$ degree.
This means the annotation is easy and robust.
We will release to public all the transformed local surface normals as an extension to
Stanford-2D-3D dataset, as well as our interactive tool for annotation.

\section{Further Analysis of Loss Functions and MultiPool Module}
In this section,
we describe further analysis on the architectural choices of where to insert
the MultiPool module, as well as new loss functions introduced in our main paper.
We conduct experiments on BSDS500~\cite{arbelaez2011contour}
and NYUv2~\cite{eigen2015predicting} datasets for boundary detection, depth and surface normal estimation,
to complement the analysis in the main paper.

\subsection{Boundary Detection on BSDS500 Dataset}
\begin{figure*}[t]
\centering
   \includegraphics[width=0.99\linewidth]{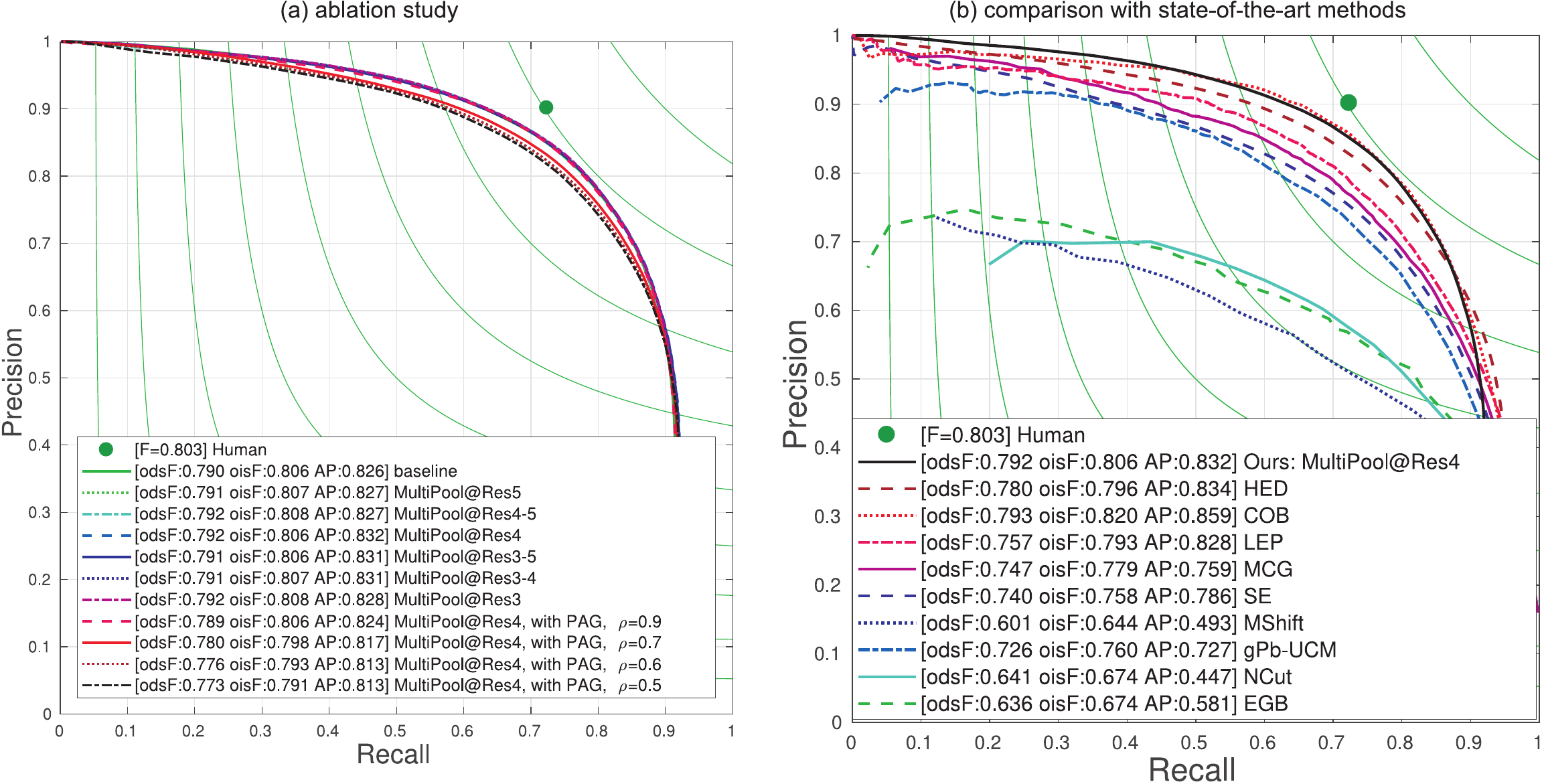}
   \caption{Precision-Recall curves on boundary detection on BSDS500 dataset.
   (a) Ablation study.
   (b) Comparison with state-of-the-art methods.
   }
\label{fig:suppl_BSDS500_boundary_curves}
\end{figure*}

In Fig.~\ref{fig:suppl_BSDS500_boundary_curves},
we show the precision-recall curves for boundary detection on BSDS500
dataset~\cite{arbelaez2011contour}.
First, Fig.~\ref{fig:suppl_BSDS500_boundary_curves} (a) summarizes our ablation study
on where to insert the MultiPool module to obtain the largest performance gain.
We observe that the best performance is achieved
when the MultiPool module is inserted at the fourth macro building block (Res4),
\ie the second last macro block or the Resnet50 architecture.
This supports our conclusion that MultiPool inserted at the second last macro building block
leads to the largest performance gain.

Moreover,
based on the ``MultiPool@Res4'' model,
we gradually insert PAG units for parsimonious inference,
and decrease the hyper-parameter $\rho$ which controls the sparsity degree of binary masks (the sparser
the masks are, the more parsimonious constraint is imposed).
In particular, note that $\rho=0.5$ means $\sim$30\% computation has been saved,
at which point our model achieves $F=0.773$, only degrading by 2.4\% performance.
This shows that the ResNet50 model has sufficient capacity for boundary detection,
and more parsimonious constraint does not harm the performance too much.
Perhaps due to this reason,
the MultiPool module does not improve performance remarkably for boundary detection.

Finally,
by comparing to state-of-the-art methods as shown in Fig.~\ref{fig:suppl_BSDS500_boundary_curves}
(b), we note our ``MultiPool@Res4'' model outperforms HED~\cite{xie2015holistically} which shares the same architecture but without
MultiPool module, and performs similarly with COB~\cite{maninis2017convolutional} which further exploits
auxiliary loss for oriented boundary detection.
This validates that the PAG-based MultiPool module improves performance by providing each pixel the
``correct'' size of pooling field.

\subsection{Monocular Depth Estimation on NYUv2 Dataset}

{
\setlength{\tabcolsep}{0.65em} 
\begin{table*}[t]
\caption{Ablation study of where to insert the MultiPool module to obtain
the largest performance gain for monocular depth estimation on NYUv2
dataset. All models are evaluated over the standard split of NYUv2 dataset,
with $L2$ loss only, and with softmax weighted average in the MultiPool module.
The performance is measured by standard threshold accuracy,
\ie the percentage of predicted pixel depths $d_i$ s.t.
$\delta=\max(\frac{d_i}{\hat d_i},\frac{\hat d_i}{d_i})<\tau$,
where $\tau=\{1.25, 1.25^2, 1.25^3\}$.
}
\centering
{
\begin{tabular}{l | c c c c c c c c c c}
\hline
metrics   & base.&MP@Res3&MP@Res4&MP@Res5&MP@Res6&MP@Res3-4&MP@Res5-6 \\
\hline
$1.25$    & 0.711& 0.721 & 0.726 & {\bf 0.737} & 0.725 & 0.726 & 0.726 \\
$1.25^{2}$& 0.932& 0.935 & 0.939 & {\bf 0.939} & 0.936 & 0.938 & 0.939 \\
$1.25^{3}$& 0.985& 0.986 & 0.986 & {\bf 0.986} & 0.985 & 0.986 & 0.985 \\
\hline
\end{tabular}
}
\label{tab:suppl_ablation_where_MultiPool_depth_NYUv2}
\end{table*}
}

{
\setlength{\tabcolsep}{1.3em} 
\begin{table*}[t]
\caption{Ablation study of loss functions and PAG unit for monocular depth estimation on NYUv2
dataset.
Our MP@Res5 model is the base model,
unless specified, all the models are trained with softmax weighted
average in the MultiPool module.
The performance is measured by standard threshold accuracy,
\ie the percentage of predicted pixel depths $d_i$ s.t.
$\delta=\max(\frac{d_i}{\hat d_i},\frac{\hat d_i}{d_i})<\tau$,
where $\tau=\{1.25, 1.25^2, 1.25^3\}$.
}
\centering
{
\begin{tabular}{l | c c c c c c c c c c}
\hline
metrics   & $L2$ loss & $L1$ loss   & $L1$+$L2$ loss& $L1$+$L2$ loss (PAG) \\
\hline
$1.25$    & 0.737   & 0.743     & 0.745     & {\bf 0.751}  \\
$1.25^{2}$& 0.939   & 0.942     & 0.944     & {\bf 0.944}  \\
$1.25^{3}$& 0.986   & 0.987     & 0.988     & {\bf 0.988}  \\
\hline
\end{tabular}
}
\label{tab:suppl_ablation_loss_depth_NYUv2}
\end{table*}
}

We provide complementary ablation study on the task of monocular depth estimation
on NYUv2 dataset.

First,
we study where to insert the MultiPool module to obtain the largest performance gain.
We train our base model using $L2$ loss function only,
and insert the MultiPool module (without PAG but the softmax weighted average operation)
at each macro building block one by one.
We list the performance of these models in
Table~\ref{tab:suppl_ablation_where_MultiPool_depth_NYUv2}.
From the table, we observe that no matter where to insert the MultiPool module,
it consistently improves the performance;
while when MultiPool module is inserted at Res5, which is the second last macro
building block,
we obtain the largest performance gain.
These observations, along with what reported in the main paper,
support our conclusion that one is able to get the best performance
when inserting the MultiPool module at the second last macro building block of a ResNet model.

Then,
we study the loss function mixing $L1$ and $L2$ as introduced in the main paper.
We train the models with MultiPool module inserted at the fifth macro block (MP@Res5),
using different loss functions,
and report the results in Table~\ref{tab:suppl_ablation_loss_depth_NYUv2}.
It's clear to see the mixed $L1$ and $L2$ loss leads to the best performance,
especially on the metric of $<1.25$, focusing on
the range of small prediction errors where $L2$ loss alone is unable to provide
a meaningful gradient.
Moreover,
we compare the model between PAG-based MultiPool and that based on softmax weighted
average operation in Table~\ref{tab:suppl_ablation_loss_depth_NYUv2}.
Again, our PAG-based MultiPool not only outperforms the weighted-average MultiPool,
but also maintains computation as well as memory storage because of the perforated
convolution (selecting spatial locations to compute).

\subsection{Surface Normal Estimation on NYUv2 Dataset}
{
\setlength{\tabcolsep}{0.4em} 
\begin{table*}[t]
\caption{Ablation study of the loss functions for surface normal estimation over
NYUv2 dataset.
Performance is measured
by mean angular error (ang. err.) and
the portion of prediction error within $t^{\circ}$ degree
where $t=\{11.25, 22.50, 30.00\}$.
Smaller ang. err. means better performance as marked by $\downarrow$.}
\centering
{
\begin{tabular}{l | c c c c c c c c c c}
\hline
metrics & cosine distance ($-\n^T {\hat\n}$)& inverse cosine ($\cos^{-1}\n^T {\hat\n}$)
    & cosine and inverse cosine\\
\hline
ang. err.$\downarrow$ & 23.3462 & 23.1191 & {\bf 22.7170} \\
\hline \hline
$11.25^{\circ}$ & 0.3163 & 0.3279 & {\bf 0.3382} \\
$22.50^{\circ}$ & 0.5995 & 0.6093 & {\bf 0.6195} \\
$30.00^{\circ}$ & 0.7240 & 0.7302 & {\bf 0.7383} \\
\hline
\end{tabular}
}
\label{tab:suppl_ablation_loss_normal}
\end{table*}
}

{
\setlength{\tabcolsep}{0.57em} 
\begin{table*}[t]
\caption{Ablation study of at which layer to insert \emph{MultiPool} module (with
softmax weighted average, w-Avg.) for surface normal estimation on NYUv2 dataset.
Performance is measured
by mean angular error (ang. err.) and
the portion of prediction error within $t^{\circ}$ degree
where $t=\{11.25, 22.50, 30.00\}$.
Smaller ang. err. means better performance as marked by $\downarrow$.}
\centering
{
\begin{tabular}{l | c c c c c c c c c c}
\hline
metrics & base. & MP@Res3 & MP@Res4 & MP@Res5 & MP@Res6 & MP@Res3-4 & MP@Res3-5     \\
\hline
ang. err.$\downarrow$
                &22.7170 &21.9951 &22.5506 &\textbf{21.9556}&22.1661&22.5183&22.5051\\
\hline \hline
$11.25^{\circ}$ & 0.3382 & 0.3560 & 0.3366 & \textbf{0.3567}& 0.3514& 0.3375& 0.3389\\
$22.50^{\circ}$ & 0.6195 & 0.6362 & 0.6188 & \textbf{0.6374}& 0.6323& 0.6198& 0.6209\\
$30.00^{\circ}$ & 0.7383 & 0.7514 & 0.7386 & \textbf{0.7526}& 0.7482& 0.7392& 0.7394\\
\hline
\end{tabular}
}
\label{tab:suppl_ablation_where_multipool_normal}
\end{table*}
}

{
\setlength{\tabcolsep}{0.3em} 
\begin{table*}[t]
\caption{Comparison of MultiPool module with PAG and softmax weighted average
(w-Avg.) over surface normal estimation on NYUv2 dataset.
Performance is measured
by mean angular error (ang. err.) and
the portion of prediction error within $t^{\circ}$ degree
where $t=\{11.25, 22.50, 30.00\}$.
Smaller ang. err. means better performance as marked by $\downarrow$.}
\centering
{
\begin{tabular}{l | c c c c c c c c c c}
\hline
metrics & MP@Res3 (w-Avg.) & MP@Res3 (PAG) & MP@Res5 (w-Avg.) & MP@Res5 (PAG)  \\
\hline
ang. err.$\downarrow$
                &21.9951 &21.9793 &21.9556&21.9226 \\
\hline \hline
$11.25^{\circ}$ & 0.3560 & 0.3591 &0.3567 & 0.3587 \\
$22.50^{\circ}$ & 0.6362 & 0.6396 &0.6374 & 0.6384 \\
$30.00^{\circ}$ & 0.7514 & 0.7523 &0.7526 & 0.7532 \\
\hline
\end{tabular}
}
\label{tab:suppl_ablation_multipool_PAG}
\end{table*}
}

Similar to the ablation study for monocular depth estimation task,
we study firstly how the proposed loss function improves performance,
then where to insert the MultiPool module for the best performance,
and lastly performance comparison between PAG-based and weighted-average
MultiPool.

In Table~\ref{tab:suppl_ablation_loss_normal},
we compare the results from models trained with different loss functions.
We can see the combination of cosine distance loss and the inverse cosine loss
achieves the best performance.
From the table,
we clearly see that the improvement on metric $11.25^{\circ}$ is more remarkable,
which focuses on small prediction errors.
This is because,
as analyzed in the main paper,
the combined loss function provides meaningful gradient ``everywhere'',
whereas the cosine distance loss alone has ``vanishing gradient'' issue when
the prediction errors become small.

We then study where to insert the MultiPool module to get the best performance in
Table~\ref{tab:suppl_ablation_where_multipool_normal}.
Note that, in this ablation study,
we didn't use PAG for binary masks,
but instead using weighted average based on softmax operator.
Consistent to previous discovery,
when inserting MultiPool at the second last macro block, we achieve the best performance.
In Table~\ref{tab:suppl_ablation_multipool_PAG},
we compare the results with PAG-based MultiPool and weighted-average MultiPool.
We conclude with consistent observation that PAG unit does not harm the performance
compared to the softmax weighted average fusion,
but instead achieves better performance with the same computation overhead,
thanks to the perforated convolution.

\section{More Qualitative Visualization}
In this section,
we visualize more results of boundary detection, semantic segmentation,
monocular depth estimation and surface normal estimation,
over the four datasets used in the paper,
BSDS500 (Fig.~\ref{fig:BSDS500_boundary_v3}),
NYUv2 (Fig. \ref{fig:NYUv2}),
Stanford-2D-3D (Fig. \ref{fig:Stanford2D3D}) and
Cityscapes (Fig. \ref{fig:Cityscapes}).
In the figures,
we show the ponder map for each macro building block,
as well as the overall ponder map.
From these ponder maps,
we can see our model learns to dynamically allocate computation at
different spatial location,
primarily expending more computation on the regions/pixels
which are regions with sharp changes, e.g. boundary
between semantic segments, regions between two depth layer, locations
around normal
changes like between wall and ceiling.
We also show all the binary maps produced by PAG units in Fig.~\ref{fig:Stanford2D3D_allMaps}
over a random image from Stanford2D3D dataset for semantic segmentation,
surface normal estimation and depth estimation.

In Fig.~\ref{fig:perforatedCNN_masks},
we visualize the learned binary masks by PerforatedCNN~\cite{figurnov2016perforatedcnns}
on NYUv2 dataset for semantic segmentation.
We also accumulate all the binary masks towards the ponder map, from which
we can see that the active pixels largely follow uniform distribution.
This is different from what reported in~\cite{figurnov2016perforatedcnns} that
the masks mainly highlight central region in image classification, which is due to the fact
that images for the classification task mainly contain object in the central region;
whereas for scene images, it is hard for PerforatedCNN to focus on any
specific location of the image.
Note again that PerforatedCNN does not support either dynamic pixel routing or fully convolutional computation,
requiring that the input image have a fixed size in order to learn fixed computation paths over the
image. In contrast, our method is fully convolutional that is able to take as
input images of arbitrary size and perform computing with input-dependent
dynamic paths.

\begin{figure*}[t]
\centering
   \includegraphics[width=0.99\linewidth]{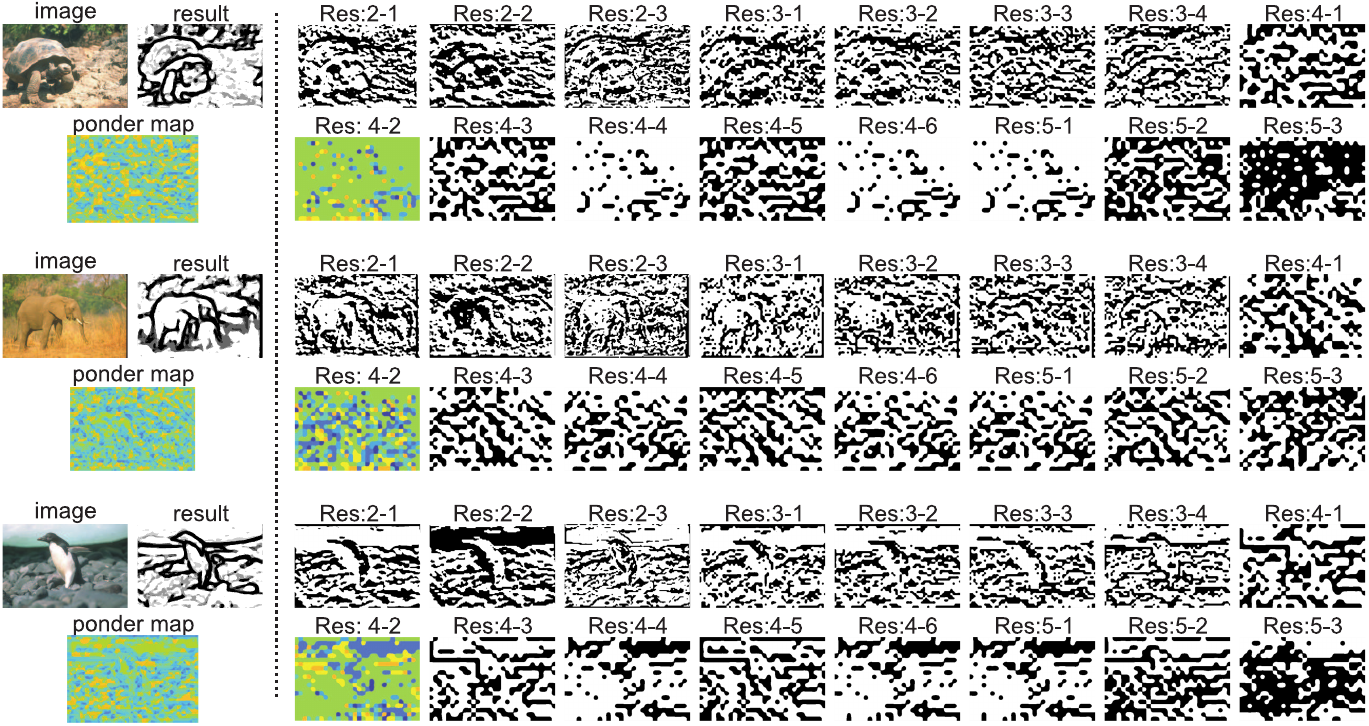}
   \caption{ Visualization on BSDS500 dataset~\cite{arbelaez2011contour}
    of sparse binary attention maps at each layer for boundary detection,
    together with the output and ponder map accumulating all binary maps.
    PAG-based MultiPool module is inserted at layer Res4-2, which is not included in
    the ponder map.
   }
\label{fig:BSDS500_boundary_v3}
\end{figure*}

\begin{figure*}[t]
\centering
   \includegraphics[width=0.99\linewidth]{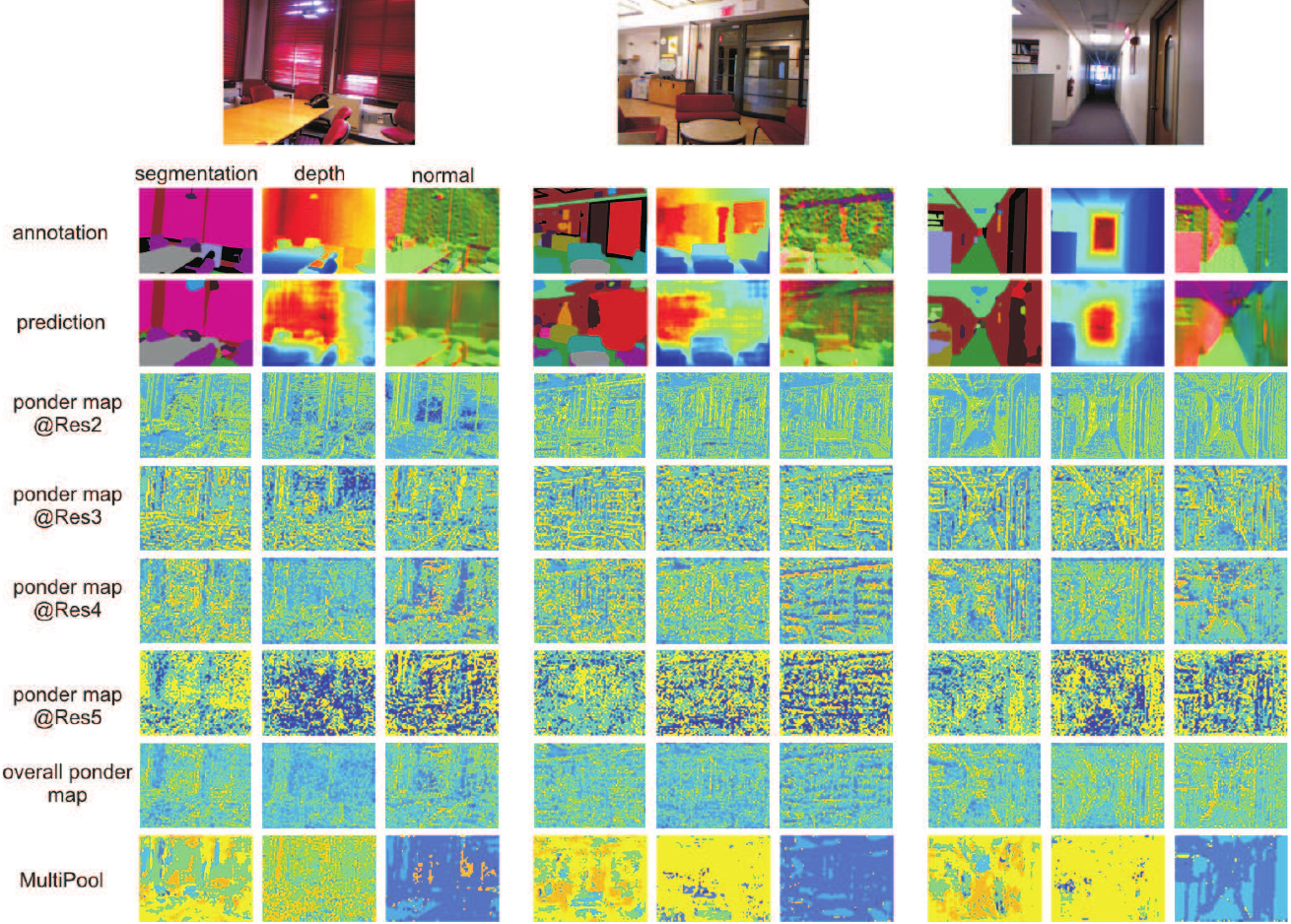}
   \caption{Visualization on NYUv2 dataset~\cite{eigen2015predicting}
   for semantic segmentation,
   depth estimation and surface normal estimation.
   Besides the overall ponder map,
   we also show the partial ponder map for each macro residual block by summing
   the sparse binary attentional maps.
   The MultiPool binary masks are not included in the ponder maps.
   }
\label{fig:NYUv2}
\end{figure*}

\begin{figure*}[t]
\centering
   \includegraphics[width=0.99\linewidth]{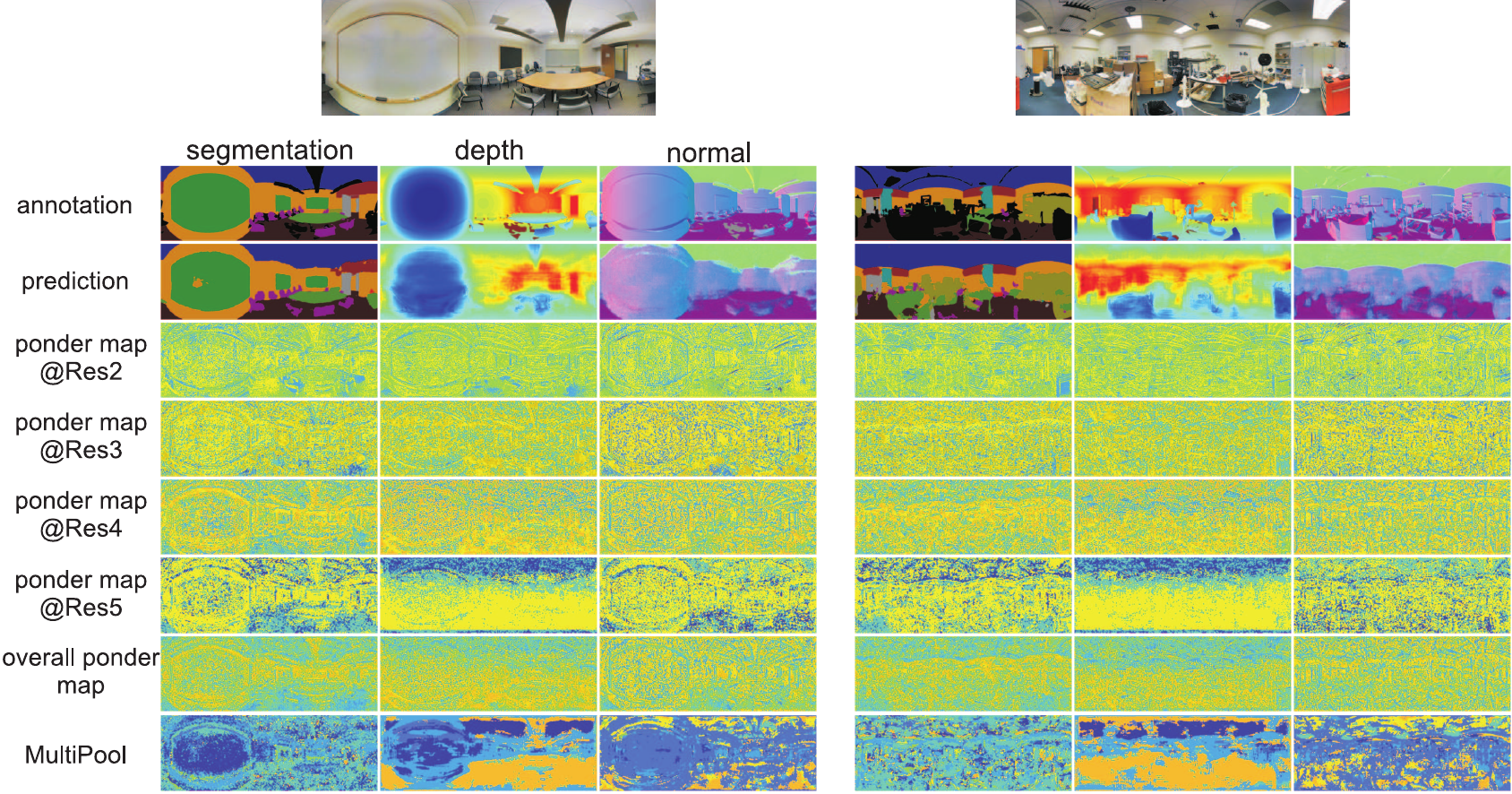}
   \caption{Visualization on Stanford2D3D~\cite{armeni2017joint} for semantic segmentation,
   depth estimation and surface normal estimation.
   Besides the overall ponder map,
   we also show the partial ponder map for each macro residual block by summing
   the sparse binary attentional maps.
   The MultiPool binary masks are not included in the ponder maps.
   }
\label{fig:Stanford2D3D}
\end{figure*}

\begin{figure*}[t]
\centering
   \includegraphics[width=0.99\linewidth]{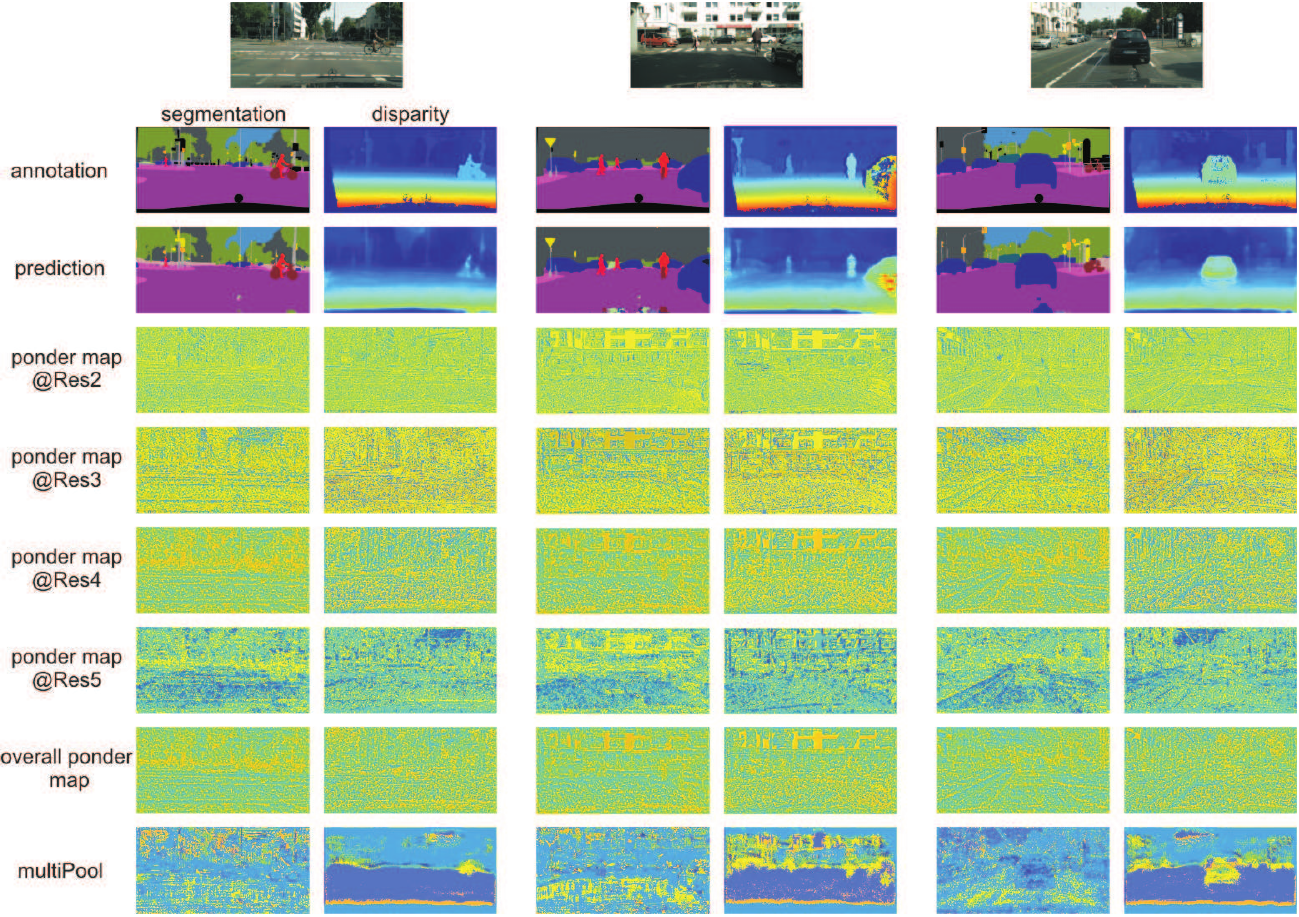}
   \caption{Visualization on Cityscapes dataset~\cite{cordts2016cityscapes}
   for semantic segmentation and
   depth estimation.
   Besides the overall ponder map,
   we also show the partial ponder map for each macro residual block by summing
   the sparse binary attentional maps.
   The MultiPool binary masks are not included in the ponder maps.
   }
\label{fig:Cityscapes}
\end{figure*}

\begin{figure*}[t]
\centering
   \includegraphics[width=0.99\linewidth]{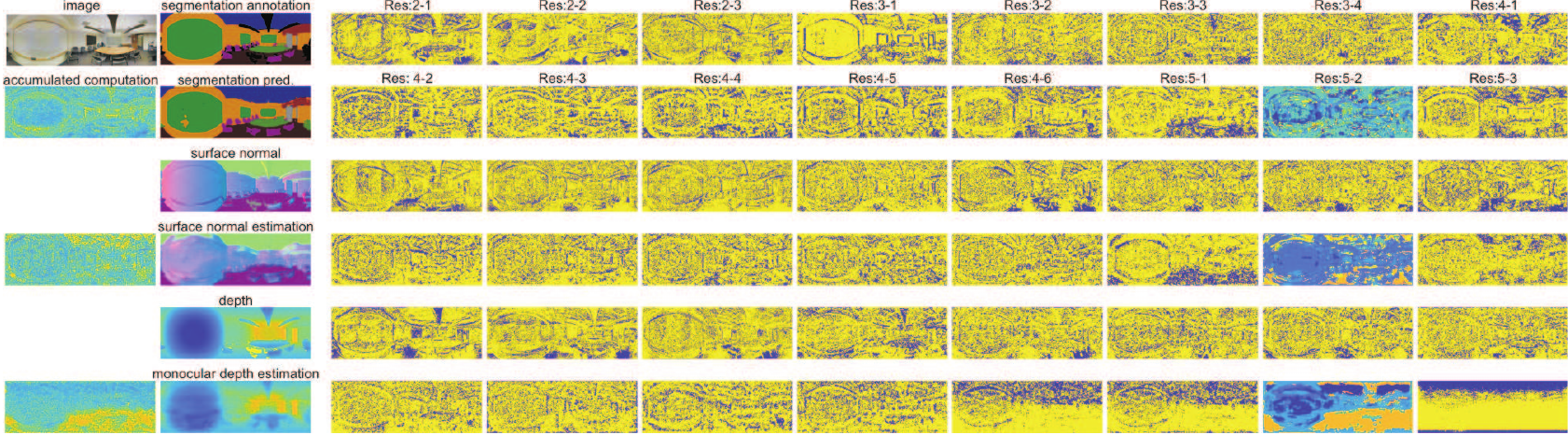}
   \caption{Visualization on Stanford2D3D~\cite{armeni2017joint} for semantic segmentation,
   surface normal estimation and depth estimation.
   Besides the overall ponder map (accumulated computation),
   we show all the binary maps produced by PAG, as well as the one in the MultiPool module at layer 5-2.
   }
\label{fig:Stanford2D3D_allMaps}
\end{figure*}

\begin{figure*}[t]
\centering
\includegraphics[width=0.99\linewidth]{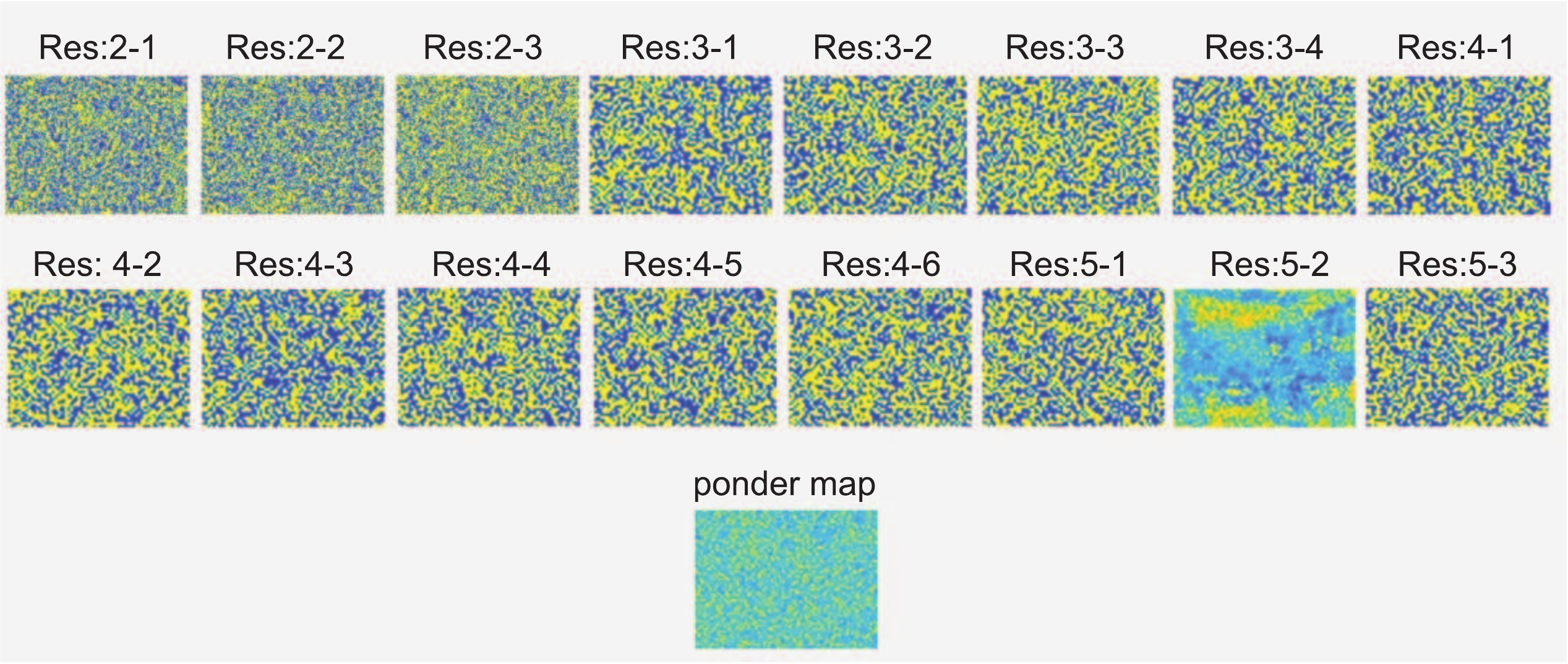}
\caption{Visualization on binary masks trained by PerforatedCNN~\cite{figurnov2016perforatedcnns}
on NYUv2 dataset for semantic segmentation.
Note that we also insert a MultiPool module at Res5-2 block. This makes it fair to
compare between our method and PerforatedCNN.
We also accumulate all the binary masks towards the ponder map, from which
we can see that the active pixels largely follow uniform distribution.
This is different from what reported in~\cite{figurnov2016perforatedcnns} that
the masks mainly highlight central region in image classification, which is due to the fact
that images for the classification task mainly contain object in the central region;
whereas for scene images, it is hard for PerforatedCNN to focus on any
specific location of the image.
}
\label{fig:perforatedCNN_masks}
\end{figure*}

\end{document}